\documentclass[pdflatex,sn-mathphys-num]{sn-jnl}
\usepackage[utf8]{inputenc}

\usepackage{graphicx}%
\usepackage{multirow}%
\usepackage{amsmath,amssymb,amsfonts}%
\usepackage{amsthm}%
\usepackage[scr=rsfs]{mathalfa}%
\usepackage[title]{appendix}%
\usepackage{xcolor}%
\usepackage{textcomp}%
\usepackage{manyfoot}%
\usepackage{booktabs}%
\usepackage{algorithm}%
\usepackage{algorithmicx}%
\usepackage{algpseudocode}%
\usepackage{listings}%
\usepackage{float}

\theoremstyle{thmstyleone}%
\theoremstyle{thmstyletwo}%

\theoremstyle{thmstylethree}%
\begin{document}

\title[Article Title]{\vspace{-10pt}{Visual-textual Dermatoglyphic Animal Biometrics: A First Case Study on~\textit{Panthera tigris}}}


\author*[1]{\fnm{Wenshuo} \sur{Li}}\email{wenshuo.li@bristol.ac.uk}

\author[1]{\fnm{Majid} \sur{Mirmehdi}}\email{m.mirmehdi@bristol.ac.uk}

\author[1]{\fnm{Tilo}\sur{Burghardt}}\email{tilo@cs.bris.ac.uk}

\affil[1]{\orgdiv{School of Computer Science}, \orgname{University of Bristol}, \orgaddress{ \city{Bristol}, \country{UK}}}

     \abstract{{Biologists have long combined photographs and sketches with textual field notes to re-identify (Re-ID) individual animals based on phenotypic traits. Contemporary AI-supported camera trapping builds on this visual tradition, with computer vision tools now supporting animal re-identification across numerous species with distinctive morphological features, such as unique stripe-like coats (e.g., zebras, tigers). Here, we present the first study to extend image-only Re-ID methodologies by incorporating precise dermatoglyphic textual descriptors -- an approach widely used in forensic science but hitherto unexploited in ecology. We demonstrate that these specialist semantics can abstract and encode both the topology and structure of animal stripe coats using fully human-interpretable language tags. Drawing on 84,264~manually labelled minutiae features across 3,355~images of 185~individual tigers \textit{(Panthera tigris)}, we provide a comprehensive quantitative case study evaluating this new visual–textual methodology in detail, revealing novel capabilities for cross-modal dermatoglyphic animal identity retrieval. To optimise performance we synthesise 2,000 ‘virtual individuals’ multimodally, each comprising dozens of life-like tiger visuals in ecologically credible camera trap configurations paired with matching dermatoglyphic text descriptors of visible coat features. Benchmarking against real-world camera trap Re-ID scenarios shows that such augmentation can double AI accuracy in cross-modal retrieval while alleviating challenges of data scarcity for rare species and expert annotation bottlenecks. We conclude that dermatoglyphic language-guided animal biometrics can overcome key limitations of vision-only solutions and, for the first time, enable textual-to-visual identity recovery underpinned by human-verifiable minutiae matchings. Conceptually, this represents a significant advance towards explainability in Re-ID and a broader language-driven unification of descriptive modalities in AI-based ecological monitoring.}}




\maketitle
\section{Introduction}\label{sec1}

{\textbf{Motivation and Context.} 
Ecological modelling, as well as wildlife conservation, hinge on the ability to monitor animals across space and time, informing a wide spectrum of critical activities including biodiversity assessments, population ecology, behavioural science, biogeography, species conservation, and related anti-poaching interventions~\cite{kuhl2013animal,chen2021data,tuia2022perspectives}. Traditionally, such efforts have relied on textual field notes and sketches, and later on photographs and invasive marking (e.g., GPS tags). Physical tagging in particular introduces unavoidable behavioural biases, higher operational costs, and sometimes safety risks to both animals and researchers due to the need of physical interaction~\cite{hebblewhite2010distinguishing,walker2011review,kays2015terrestrial}. Overcoming some aspects of this, recent advances in computer vision have now boosted the effectiveness of non-invasive alternatives able to scale photographic techniques through widely automated species detection~\cite{beery2019efficient,leorna2022human,gadot2024crop}. However, distinguishing individuals reliably based on only sparse, unconstrained real-world imagery of often rare species remains an ongoing research challenge~\cite{moskvyak2021robust,schneider2022similarity}. In fact, a recent horizon scan~\cite{reynolds2025potential} identified individual animal identification as one key priority for further developments in conservation technology.}

{Among the many morphological and behavioural features that distinguish individuals, fur stripes in mammals akin to Turing patterns~\cite{kondo2010reaction,eizirik2010defining}---such as the integument of tigers \textit{(Panthera tigris)} or the banded pelage of plains zebras \textit{(Equus quagga)}---are most prominently accessible in visual recordings. The ease of their measurability combined with substantial, stable inter-individual variation within populations~\cite{kondo2002reaction} render such phenotypic coat traits robust and readily usable individual markers~\cite{burghardt2008visual,bolger2012computer,cihan2023identification}. They satisfy all essential requirements for biometric entities as set out by Jain~\cite{jain2005biometric}, that is regarding uniqueness, permanence, universality, comparability and, critically, measurability.}

{Camera trap re-identification (Re-ID) systems leverage these distinctive body surface patterns via annotated imagery to distil identity-relevant information together with metadata into neural network form that can match individuals in photographic evidence after system training. These approaches now help to perform census tasks as well as reconstructions of movement trajectories and behavioural profiles~\cite{burton2015wildlife,ravoor2020deep,ma2025deep}. However, the central bottleneck in this domain remains the limited robustness and interpretability of Re-ID algorithms and related applications.}

{Critical difficulties that underpin these shortcomings are posed by both species rarity, as well as the reliance on expensive expert-driven annotation for consistently cataloguing the
identity-bearing morphological traits of interest; be that discussed dermatoglyphic coat features, ear pinna notches in elephants, trailing edge contours of whale flukes or other traits~\cite{li2020atrw,bedetti2020system,prinsloo2021unique,petso2022review,cheeseman2022advanced,gomez2023re,meguro2024stripe,nepovinnykh2024species,gholami2025ai}. Given that modern artificial intelligence~(AI) systems require significant training data, such availability constraints significantly hinder the transition from traditional handcrafted feature-based approaches to modern end-to-end deep learning models.}

{Beyond data constraints, a multitude of additional challenges further complicate visual animal Re-ID, including pose variations, viewpoint disparities, the complexity of natural environments, and uncontrolled illumination
conditions. These factors collectively induce significant intra-class variation, which pose fundamental limitations to reliable individual animal differentiation in the wild when relying on relatively small corpora of image-based information only~\cite{moskvyak2021robust,wang2021giant,zheng2022wild,li2024adaptive,han2024multi,mulero2025addressing}.}

\begin{figure}[t]
\centering
\includegraphics[width=\textwidth]{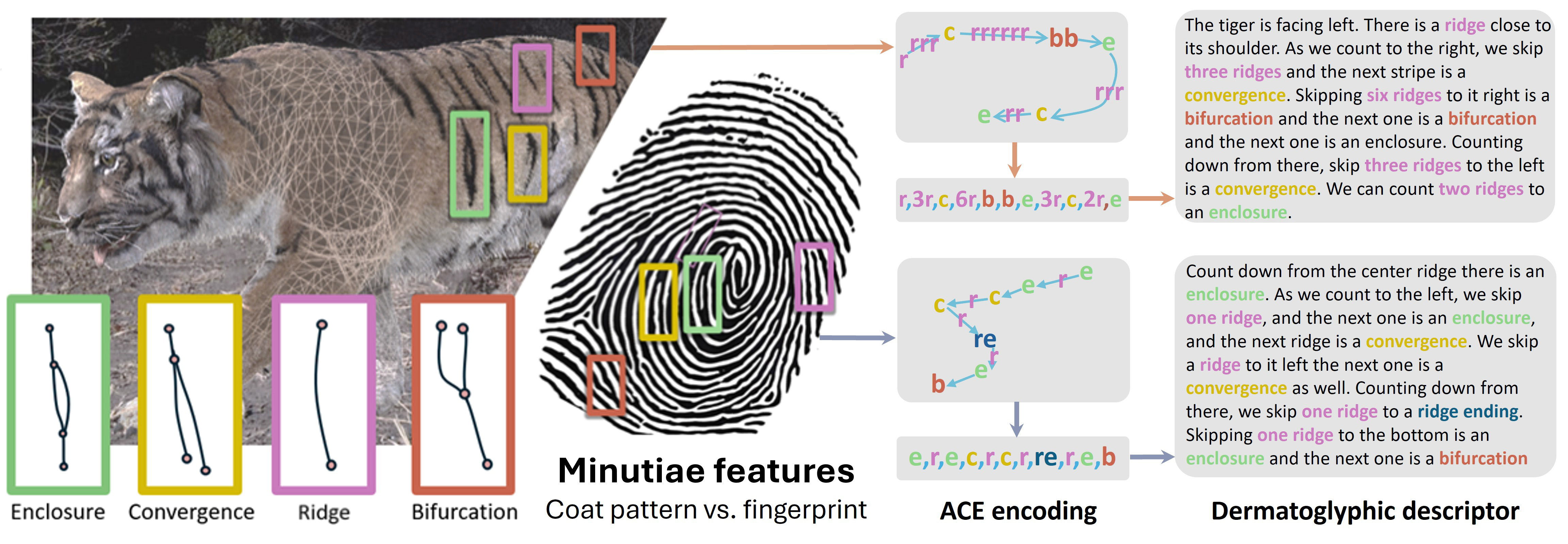}
\caption{{\textbf{From Visual Minutiae Features to Dermatoglyphic Text Descriptions.} 
Topological definitions of four common dermatoglyphic structural details~(i.e. minutiae) are shown at the bottom left, with examples of these precise structures within a fingerprint to their right, as routinely determined during  forensic analysis. Above to the left, corresponding stripe arrangements are displayed on a synthesised tiger coat pattern. Following arrows to the right, the principles of the fingerprint ACE process~\cite{stevenage2016fact,needham2022collaborative} are illustrated. Minutiae are identified along anatomically informed scan paths, sequentially encoded, and transformed into structured textual form. The resulting text is precisely interpretable, manually checkable against the visual, and at the same time compactly captures the individuality of the encoded pattern. This enables both `white box' matching and compact descriptor assembly across pattern categories beyond the scope of traditional computer vision-only models.
}}\label{minutiae}
\end{figure}

{Meanwhile, forensic science has long maintained highly refined expert terminology for dermatoglyphic analysis~(see Fig.~\ref{minutiae}), furnishing courts worldwide with interpretable and evidentiary descriptions of human skin impressions,  such as fingerprints or palm prints~\cite{stevenage2016fact,chauhan2017latent,dhaneshwar2021investigation,needham2022collaborative,khodadoust2024enhancing}. 
Yet, despite early work on symbolic coat pattern sequencing~\cite{petersen1972identification}, computational dermatoglyphic text encodings of animal coats have so far neither been attempted nor been used for Re-ID. }

{\textbf{Dermatoglyphic Animal Biometrics.} 
Although previous work has recognised the conceptual alignment~\cite{burghardt2008visual,allen2011leopard,kuhl2013animal,lee2018seeing,chan2019pat,staps2023development} between modern forensic terminology~\cite{yager2004fingerprint,martins2024fingerprint} and the structure of biologically distinctive stripe configurations observed in Turing-patterned coat phenotypes~\cite{stoney1986critical,ouyang1991transition}, a computational association of dermatoglyphic textual descriptors with corresponding animal visuals has never been attempted. As illustrated in Fig.~\ref{minutiae} and used throughout the paper for tigers~\textit{(Panthera tigris)}, biometric minutiae as employed in forensics, such as ridge endings and bifurcations in human fingerprints, show direct structural correspondences with analogous pelage configurations in striped animal coats. Arrangements of these minutiae on fingerprints are routinely interpreted forensically as sequential symbolic encodings via Analysis-Comparison-Evaluation~(ACE)~\cite{malhotra2020understanding,deshpande2022study} frameworks. These are well established tools that provide a conversion mechanism from visual to text underpinning today's language-based forensic reporting.}

{We re-purpose the ACE encoding approach for coat pattern characterisation; that is we employ it for the textual description of striped surface morphologies unique to individual animals~(see Fig.~\ref{minutiae}). Conceptually, the precise specialist terminology that forms the foundation of dermatoglyphic forensic science is thereby extended to the domain of animal biometrics~\cite{kuhl2013animal, tuia2022perspectives}. In contrast to systems that highlight visual correspondences of matching coat elements to get closer to explainable animal Re-ID~\cite{shrack2025pairwise,nepovinnykh2024species}, dermatoglyphic encodings bridge the semantic vision-language gap naturally. They  provide a discrete, human-readable language representation if pattern information most relevant to Re-ID. Notably, such minutiae-based ACE encodings are topological in nature and thereby widely invariant to viewpoint variations as well as surface deformations facilitating more compact and robust representations.}

{\textbf{Bridging Animal Biometric Modalities.} To operationalise this concept and bridge the semantic modality gap between vision and language descriptors for striped coat patterns, we present a scalable image-text co-synthesis framework to support network training at scale. The system is capable of generating infinite numbers of synthetic individuals described both visually~\textit{and} textually for tigers. Based on such large-scale structured data, deep learning AI architectures can be used to correlate symbolic sequences expressed in dermatoglyphic language with associated classes of visual minutiae representations in coat patterns~(see Fig.~\ref{minutiae}) enabling so far unexplored cross-modal retrieval.}

{\textbf{Visual-textual Co-synthesis in Practice.} Scalable image-text co-synthesis capable of generating synthetic individuals visually \textit{and} textually forms the core of our practical proof-of-concept system, which uses~\textit{Panthera tigris} as a model organism. We generate 24,000 different text-image exemplifications of 2,000 virtual tigers with anatomically consistent 3D pelage representations by synthesising surface textures~(UV maps) from minutiae distributions~(see Fig.~\ref{statistical pattern synthesis}) ready to be projected onto any corresponding 3D animal pose representation. Variations across virtual individuals can therefore be co-captured in both camera trap-like, synthetically rendered images as well as ACE-encoded textual descriptors. To improve the authenticity and diversity of this virtual data corpus and to closely approximate real-world camera trap visuals, we adopt the following strategies during the synthesis process:} 

{\textit{\textbf{i) Ecological Validity:}} We perform a statistical analysis of minutiae distributions across empirically documented body regions of wild tigers~(see Fig.~\ref{statistical pattern synthesis}) to drive coat creation for virtual individuals. This alignment between synthetic and observed data preserves the statistical distribution of real biological trait expression in virtual populations as to minimise the semantic gap introduced by data augmentation.}

{\textit{\textbf{ii) Coat Consistency under Deformation:}} Recognising distortion challenges introduced by UV~\cite{flavell2010uv,poranne2017autocuts,srinivasan2024nuvo} unwrapping during 3D texture mapping, we apply a non-linear deformation correction procedure based on Radial Basis Function~(RBF)~\cite{majdisova2017radial} interpolation. This method effectively simulates aspects of skin distortions and compensates for warping artifacts. It ensures accurate projection of pelage patterns onto any 3D mesh surface while maintaining structural pattern integrity~(see Fig.~\ref{rbf and skeleton} (a)).}

{\textit{\textbf{iii) Postural Repertoire:}} We embed a skeletal rig within the 3D tiger mesh and apply 
region-specific articulation constraints. This enables joint-specific, physiologically plausible deformation control across an entire range of animal postures. By tuning joint weight distributions, displacement vectors, and inter-joint mobility parameters, we simulate a realistic repertoire of natural postures observed in typical camera trap footage~(see Fig.~\ref{rbf and skeleton} (b)).}

{\textit{\textbf{iv) Realistic Fur Synthesis:}} To bridge the domain gap between synthetic and real-world data during training and inference, we utilise 3D software~\cite{naiman2017houdini,borkiewicz2019cinematic,SideFX2023Houdini} to generate biologically accurate fur on the model surface, refining key features of stripes and pattern details~(see Fig.~\ref{hair}).}

{\textit{\textbf{v) Scenic Image Fidelity:}} To account for the complexity of natural environments, we {employ} high-dynamic-range imaging~(HDRI) to simulate realistic environmental lighting and use virtual cameras to capture images from variable perspectives akin to camera trapping scenarios, increasing the visual diversity and realism of the artificial dataset. Additionally, we {integrate} deep image harmonisation techniques~\cite{Harmonizer} to seamlessly blend synthetic individuals with their backgrounds, further enhancing environmental coherence of the synthetic data~(see Fig.~\ref{render}).}

{In summary, by precisely modelling the biometrically most important traits of the species at hand down to minutiae level along an automated image–text co-synthesis pipeline (as described above), we effectively mitigate the scarcity of training data and can construct a large-scale, high-quality synthetic dataset -- where top-end generative AI tools~\cite{esser2024scaling,zhang2023adding,betker2023improving,comanici2025gemini} would struggle to maintain character consistency. This, in turn, allows for generating training data suitable for learning multi-modal and cross-modal animal re-identification capabilities directly and at scale. Most importantly, the resulting training corpus is \textit{biologically} grounded and reflects representative camera trap settings~(see Fig.~\ref{render}).}

{\textbf{Real-world System Evaluation and Experimental Setup.} 
To investigate the discriminative capacity of dermatoglyphically grounded descriptors for the species at hand, we first {curate} and systematically {annotate} a real-world dataset of 84,264 manually minutiae features in 3,355 images of 185 wild tigers based on~\cite{harihar2019tigerrajaji,li2020atrw}. This dataset specifically supports our minutiae-grounded multi-modal and cross-modal biometric AI evaluation tasks beyond traditional image-to-image Re-ID, which forms the baseline:}

{\textit{\textbf{i) Dermatoglyphic Text Re-ID:}} First, text-only re-identification experiments are performed~(see Section~\ref{ResultA}) where models are trained exclusively on ACE-based text descriptions. This is to {demonstrate} that near perfect individual differentiation is possible based on only these formal language descriptors, highlighting the semantic robustness of dermatoglyphically grounded text descriptors for the species at hand~\textit{despite} congruency of the vision-text ground truths -- that is all challenges regarding image quality, minutiae interpretability, and related content are congruently reflected in the descriptor string based on human expert annotation.}

{\textit{\textbf{ii) Multi-modal Visual-Textual Re-ID:}} Second, we quantify in how far in multi-modal learning scenarios -- where visual and textual representations are jointly encoded -- the inclusion of these lexical descriptors can carry-through enhanced model performance and outperform image-only settings~(see Section~\ref{ResultA}).}

{\textit{\textbf{iii) Cross-modal Textual-to-Visual Re-ID:}} Most importantly, we explore and evaluate \textit{novel} cross-modal text-to-image retrieval. This simulates practical conditions in ecological monitoring where either 1)~textual field records have to substitute for input photography, or 2)~a manual ACE transcription of an existing, poor quality image (e.g. image noise, partial occlusion,  view point challenges) may be required to enable identity retrieval. Leveraging our synthetic image–text training corpus, we show that such retrieval can be implemented effectively by deep networks, affirming the viability of textual encodings for animal image identification. This is underpinned by ablation studies on dataset scale as well as ACE minutiae presence. Anchor permutation studies add to this and quantify critical relationships between ID-relevant dimensions, such as descriptive format, data granularity, and data composition.}

{These experiments not only underline viability and applicability of forensically inspired and language-enhanced AI techniques for animal Re-ID, but also offer a conceptually new, cross-modal perspective for research in animal biometrics in general. Grounded in a demonstrably effective vision-language encoding strategy and supported by an efficient data synthesis pipeline, the concepts put forward establish a novel sub-domain for cross-modal machine learning, that is: dermatoglyphic visual-textual animal biometrics. By integrating methodologies from forensic science, field ecology, and computer vision, we construct a computational, human-verifiable bridging mechanism between visual coat photography and textual identity descriptions. There now follows a detailed exposition of results and associated methodology.}



\section{{Results}}\label{sec2}

\subsection{{Assembly of Dermatoglyphic Descriptors}} \label{subsec2.1}
{\textbf{Encoding Stripe Coats via the ACE Protocol.} Experiments are grounded in a minutiae-based re-purposing of the ACE framework for the symbolic characterisation of Turing-patterned animal stripe coats. Using \textit{Panthera tigris} as a model taxon, we anchor our framework in four principal minutiae types -- ridges, bifurcations, convergences, and enclosures~(see Section~\ref{method_minutiae_augmentation} for full definitions) -- that mirror essential fingerprint morphologies as seen in Fig.~\ref{minutiae} middle. These components form the elementary vision-language bridge for constructing discriminative textual identity representations. In addition, their fundamental reliance on topological rather than geometrical features renders the approach robust with regard to geometric transforms of the animal surface induced by view-point and pose, as well as non-self-occluding body articulations.}

{\textbf{Robustness of Identity Encodings.} In the original ACE framework, minutiae descriptors as well as their chaining are typically oriented using radial positioning anchored at the fingerprint core~(see Fig.~\ref{minutiae} middle) -- radiating outward. This allows for robustness against acquisition-dependent rotational variation and surface distortion. In contrast, animal bodies only offer imprecise anatomically grounded references or anchor points~(e.g., shoulder tip, base of tail). Nevertheless, a scan path may still be constructed leveraging coarse anatomical stability -- for instance, in the case of tigers, running from the shoulder to the tail and then returning towards the forelimb closing the loop~(see Fig.~\ref{minutiae} centre top). Along such an anatomically guided scan trajectory, a reproducible sequence of pelage minutiae can then be extracted across individual specimens. To account for the absence of an exact reference location to start the sequence, we train systems across an entire set of permutations of the minutiae sequence along the scan trajectory to generate robustness. Note that such permutations also mitigate directionality errors introduced by pose variation and related anchor ambiguity. Motivated by a need for human readability and utilising ridge-counting adapted from ACE for topological referencing, we translate these sequences into compacted natural language descriptors~(see Fig.~\ref{minutiae} top right for specific example). This process establishes a direct mapping between uniqueness-bearing morphological coat traits and dermatoglyphic language constructs providing a unified, human-interpretable, and highly discriminative feature space for minutiae-driven re-identification -- bridging the vision-text domain gap.}

\begin{figure}[t]
\centering
\includegraphics[width=\textwidth]{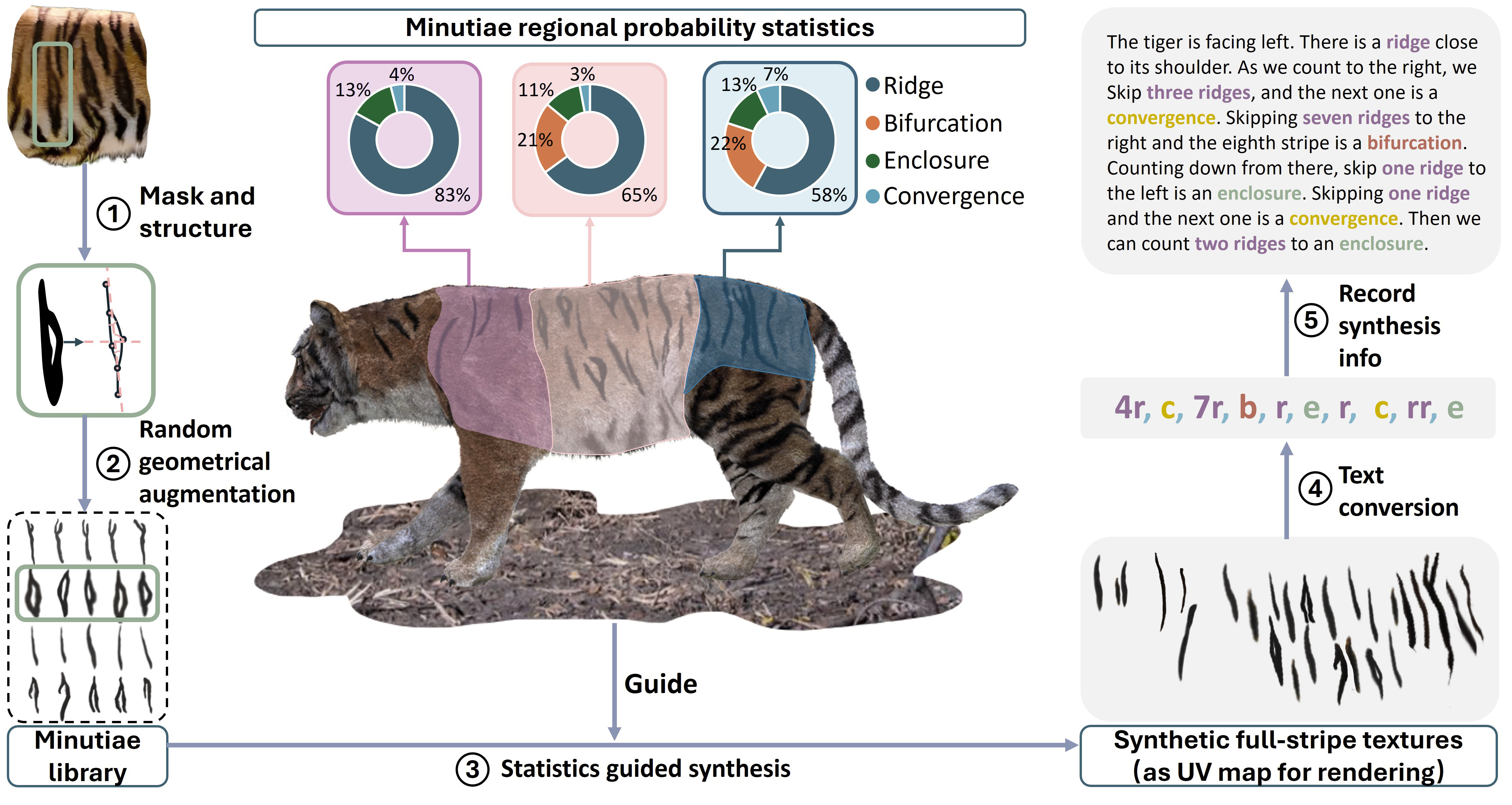}
\caption{{{\textbf{Visual-textual Co-Synthesis of Virtual Animal Coats.} 
Utilising spatial statistics of minutiae types is real anatomical pattern distributions across a population, we assemble full stripe textures based on a large minutiae library. 
The latter is constructed through keypoint augmentation, from which region-specific instances are sampled.
ACE descriptors and language descriptions are paired with their expression in full-stripe textures to, once rendered, represent a virtual individual multi-modally. 
This forms the basis for producing any number of virtually rendered animal individuals simulating camera trap scenarios in order to produce realistic imagery for cross-modal AI training.
}}}\label{statistical pattern synthesis}
\end{figure}

\subsection{{Visual-textual Co-synthesis of Virtual Individuals}}\label{subsec2.2}
{\textbf{Descriptor-driven Synthesis of Individual Coat Patterns.} To allow for AI training well beyond only measured pattern datasets and reduce any semantic disparity between synthetic and real-world animal imagery, it is imperative to generate life-like and morphologically diverse visuals of coat patterns that express realistic ranges of textually described dermatoglyphic configurations. Such a process not only mitigates the scarcity of training data, but ensures full visual–textual alignment between modalities. In order to drive the synthesis of minutiae visuals based on manifestations in real populations, we first extract distinct minutiae typologies from high-resolution imagery of actual wild~\textit{Panthera tigris} specimens. Using these as texture templates, we apply stochastic geometric augmentations to them using controlled, species-specific spatial keypoint parameters to produce a broad variety of biologically feasible visuals. Augmentations used include random translation, affine distortion, local warping, rotation, and scaling within biologically plausible parameters as detailed in Section~\ref{method_minutiae_augmentation}.}

\begin{figure}[t]
\centering
\includegraphics[width=\textwidth]{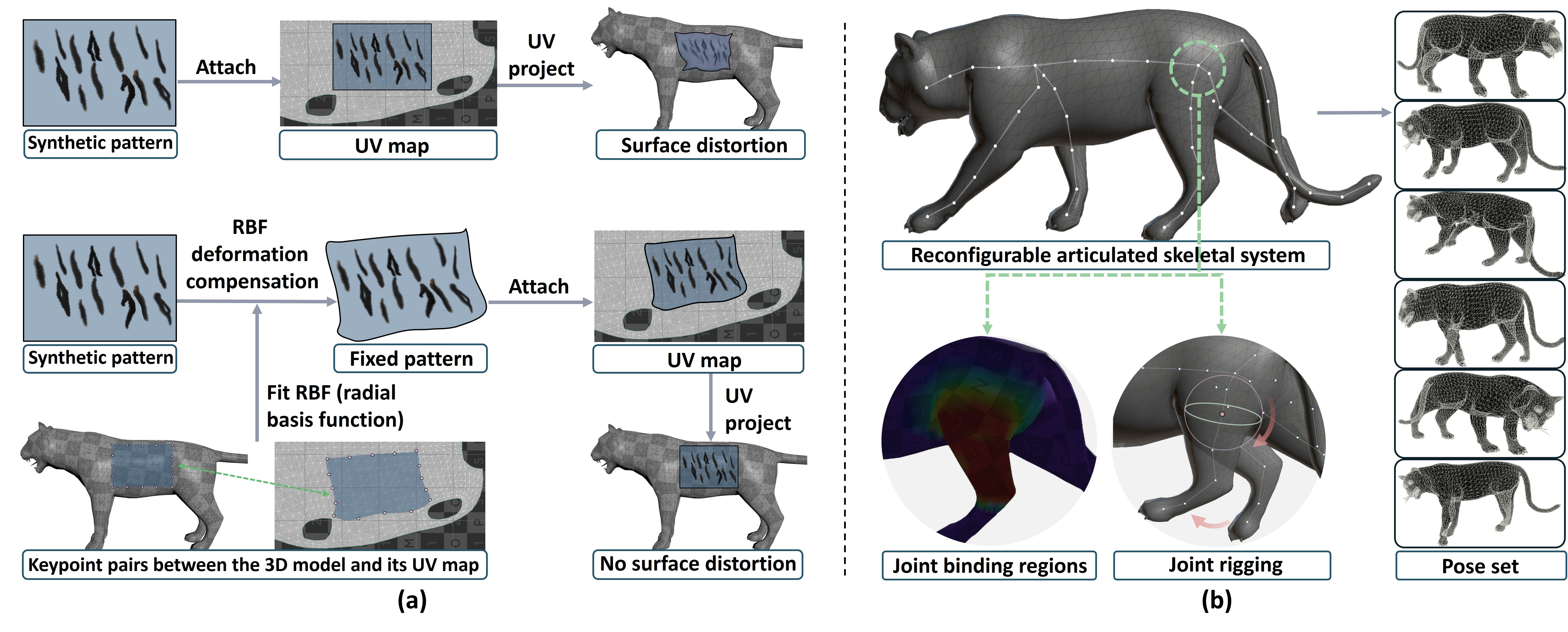}
\caption{{\textbf{Distortion-corrected Texture Mapping and 3D Pose Modelling.} 
\textbf{(a)}: Texture distortion caused by non-linear UV projections is corrected using RBF approximation, preserving anatomical consistency of the pelage pattern. \textbf{(b)}: A controllable skeletal system with region-specific bindings enables natural 3D pose modelling by adjusting joints according to camera trap references. Examples of resulting animal poses that are realistically observable in real-world camera trap settings are shown on the right.}}\label{rbf and skeleton}
\end{figure}

{The resulting library (database) of diverse minutiae manifestations for the species at hand forms the basis for subsequently assembling full-stripe texture maps for each virtual~\textit{Panthera  tigris} individual~(see Fig.~\ref{statistical pattern synthesis} left) taking the associated ACE sequence as a parameter. To preserve anatomical realism in spatial pattern distributions during ACE sequence generation for these assemblies, we partition the tiger’s body into three discrete anatomical regions and calculate region-specific 
minutiae occurrence probabilities based on empirical analysis of 711 minutiae of 20 real specimen~(see Fig.~\ref{statistical pattern synthesis} middle). These probabilistic models inform minutiae placement during texture synthesis, yielding biologically coherent and spatially consistent pattern layouts.  This synthesis pipeline  encodes the spatial location, 
type, and parametrisation of each minutiae element as explicit text with the visual dataset -- enabling precise multi-modal biometric consistency~(see Fig.~\ref{statistical pattern synthesis} right). Full-stripe textures can then be produced as UV maps and be projected onto 3D animal mesh needed to render life-like virtual individuals~(see Fig.~\ref{rbf and skeleton} (a)). }

{\textbf{Distortion-compensated Projection of Coat Patterns onto 3D Mesh.} Mapping coat textures from 2D UV space onto 3D animal meshes introduces spatial distortions due to non-linear surface stitching between different elements of the associated UV space. These non-linearities prevent the accurate alignment of associated texture patches on the 3D surfaces of virtual target individuals -- essentially breaking visual stripe continuity. In most domains, human animators would correct for such mismatches manually to avoid warped or misaligned markings that compromise morphological fidelity. For our application of automated virtual individual generation at scale, we instead employ a non-linear UV correction procedure using radial basis function~(RBF)~\cite{majdisova2017radial} deformation modelling~(see Fig.~\ref{rbf and skeleton} (a)). By densely sampling homologous point correspondences between the surface mesh and its UV flattening, we fit an RBF-based transformation to capture the local deformation fields as given in full detail in Section~\ref{method_RBF_and_rendering_detail}.
The resulting inverse deformation is applied to synthesised textures, ensuring that projected stripe patterns are mapped without misalignment and retain anatomical consistency across the body surface. By integrating minutiae synthesis with deformation correction, we achieve high-fidelity pelage pattern rendering suitable for downstream deep learning and biometric analysis.}

{
\textbf{Anatomical Pose and Articulation of 3D Models.} To enable a wide range of biologically plausible postures, we integrate a fully controllable skeletal rig within a high-fidelity 3D representation of \textit{Panthera tigris}. The embedded skeletal framework comprises anatomically defined segments—including the head, cervical vertebrae, torso, limbs, and tail—allowing realistic simulation of species-specific locomotor patterns (see Fig.~\ref{rbf and skeleton} (b)). Joint articulation is implemented using region-based control mechanisms, wherein each skeletal joint is assigned a binding domain with realistically calibrated deformation weights, displacement parameters, and linkage constraints. This approach mitigates geometric artefacts such as surface collapse or mesh folding during pose transitions, 
ensuring visual realism and consistency with real imagery of the species. To capture a representative range of behavioural postures in camera trapping scenarios, we analyse a large corpus of field images and identify key motion states including quadrupedal stance, ambulatory stride, cervical rotation, axial torsion, cranial lowering, and elevation. These pose archetypes are then encoded within the skeletal system and support 
biologically valid pose synthesis frequently observed in camera trap imagery.}

\begin{figure}[htbp]
\centering
\includegraphics[width=\textwidth]{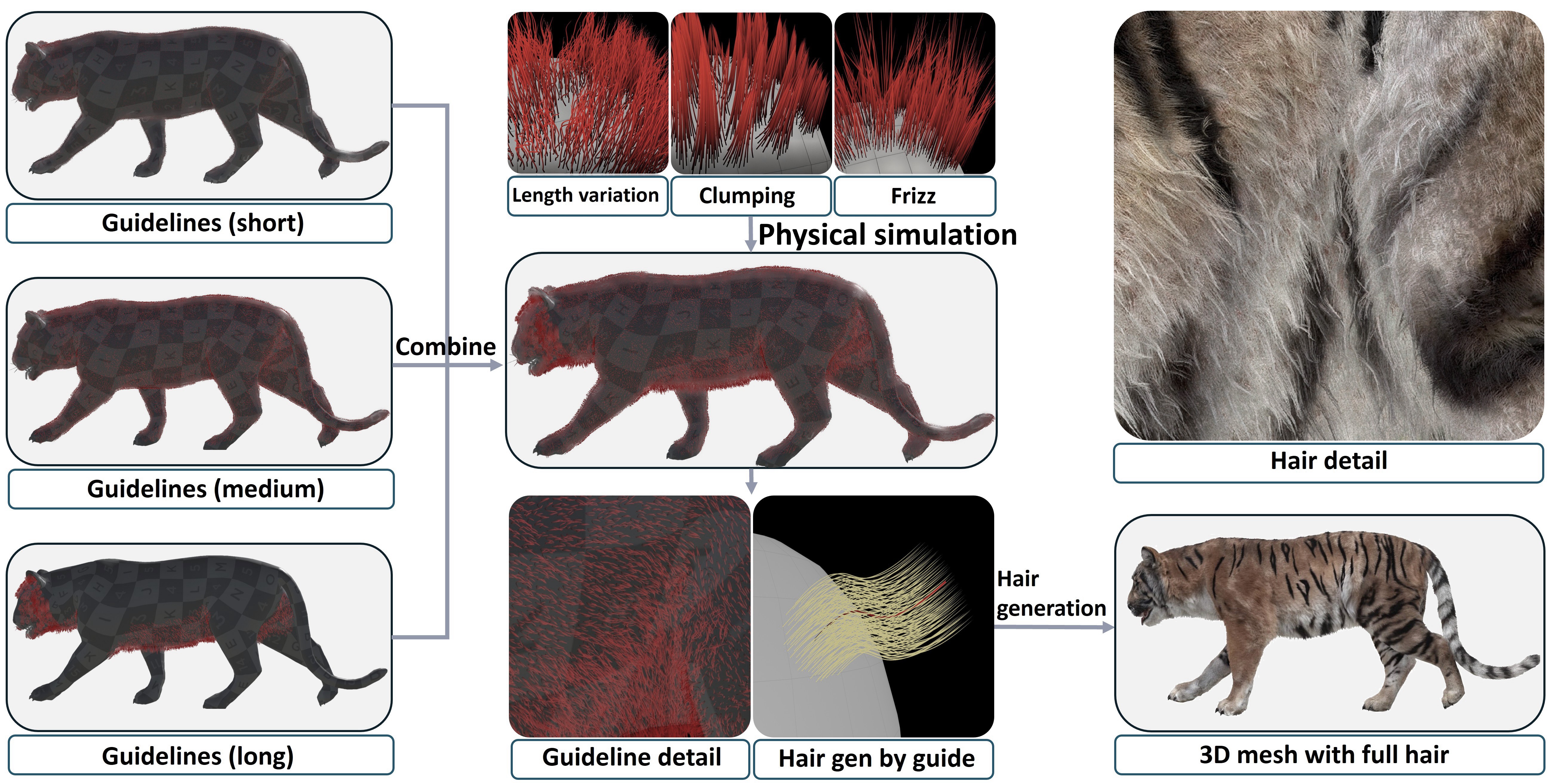}
\caption{{\textbf{Biomimetic Pelage Synthesis.} 
Anatomically-driven guides on the model surface are segmented by hair length~(short, medium, long) across anatomical regions; fur is generated along these guides, with pelage shapes driven by orientation and length parameters, and physical simulations achieved through perturbation, clumping, and frizz. The resulting fur textures~(right) have high fidelity and reduce any semantic gap to real population photography otherwise impeding on network training.}
}\label{hair}
\end{figure}

{\textbf{Biomimetic Pelage Synthesis.} To improve realism of coat renderings further we use advanced graphics techniques. We manually define guide curves for hair across the surface topology of the 3D mesh to control strand orientation, length distribution, and follicular density with region-specific, biologically plausible fidelity~(see Fig.~\ref{hair}). These spline-based guides simulate anisotropic fur growth per anatomical zone. Pelage primitives are classified into short, medium-long, and long strand sets, which correspond to the general body surface, facial and trunk segments, and specialised areas such as the cheeks and abdomen. Individual strands are instantiated along guide paths using strand-based geometry generation, with local strand behaviour modulated by parameter fields and iterative relaxation steps that emulate biomechanical properties and yield life-like mammalian fur visuals~(see Fig.~\ref{hair}). To achieve this, we apply a suite of physics-based simulations including stochastic variation in strand length, localised clumping algorithms for cohesive fibre grouping, and frizz models that capture curvature induced by self-contact and environmental perturbation. Material maps defined at the follicular root level govern albedo and pigment distribution, while tip regions incorporate light transmission and light interaction under variable illumination~(see Section~\ref{method_RBF_and_rendering_detail} for full modelling details). Overall, this integrated strategy, spanning anatomical pose control and biologically hair modelling, ensures that synthesised animals provide a robust foundation for generating high-quality visuals for neural network training.}

\begin{figure}[htbp]
\centering
\includegraphics[width=\textwidth]{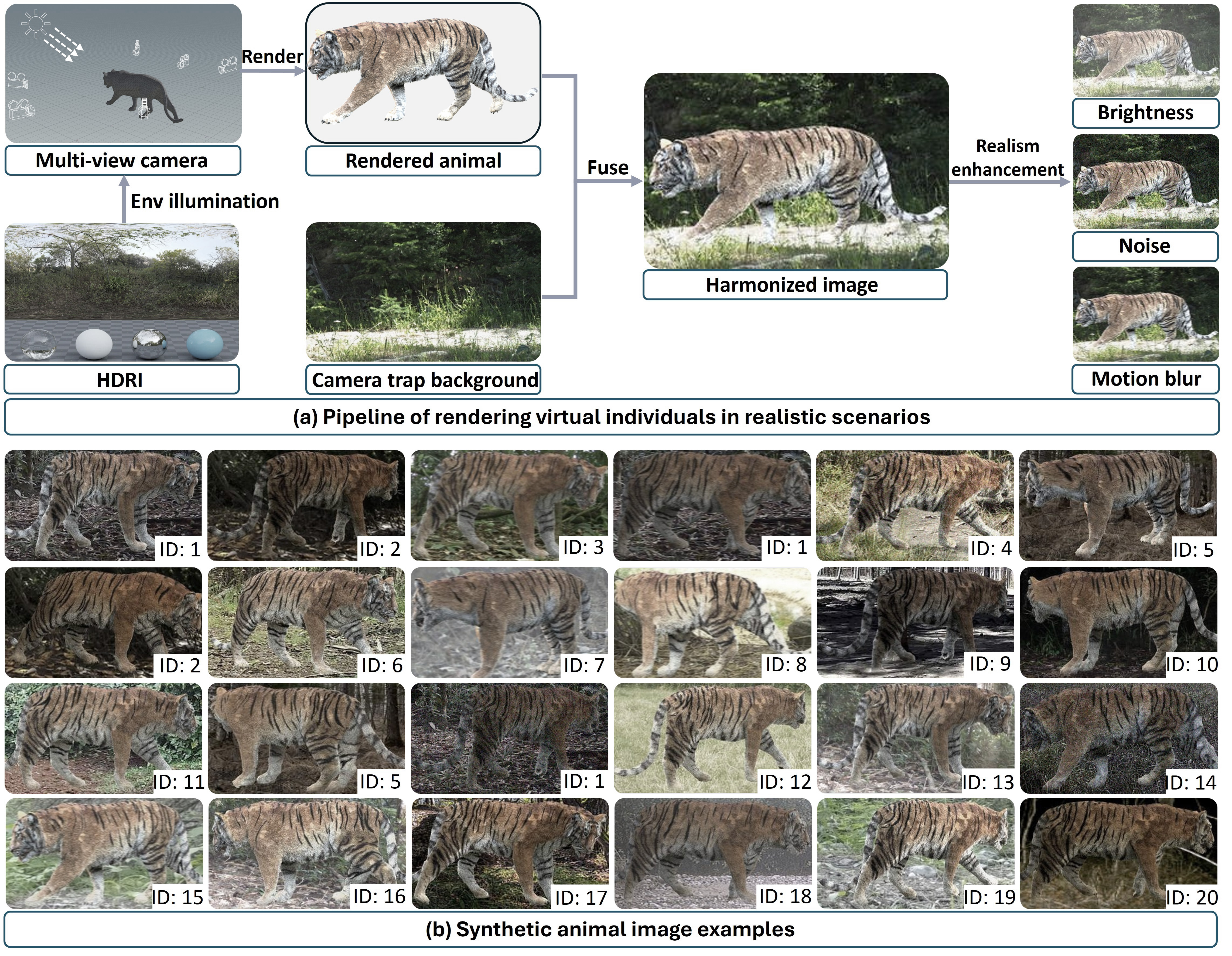}\vspace{-10pt}
\caption{{\textbf{Visual Synthesis of 24,000 Camera Trap Images of Virtual Tiger Identities. } 
\textbf{(a)}:~Where current model-free AI synthesis would struggle to generate virtual individuals with truly matching image-text pairs, our automated virtual image construction pipeline produces pattern-consistent visualisations of life-like (virtual) tigers in camera trap scenarios. Virtual cameras capture 3D animal meshes with distinct coat pattern identity from multiple, task-specific angles under realistic HDRI lighting, with backgrounds from real camera trap images, and {are} post-processed via harmonisation and augmentation, mimicking field conditions {to reduce} the semantic gap to real-world imagery. \textbf{(b)}: 24 samples within resolution range 301x156 to 413x195 from the 24,000 resulting multi-view animal renderings used for system training labelled with corresponding virtual animal IDs and exemplifying variations in real-world illumination and viewpoints akin to camera trapping. Note that identities are assigned -- as in field protocols in use~\cite{li2020atrw} -- to single sides~(left or right) of the tiger. The reader may confirm virtual identities themselves to experience the difficulty of Re-ID from images for the species at hand.}}\label{render}
\end{figure}

\subsection{{Rendering Virtual Individuals in Realistic Scenarios}}\label{subsec2.3}

{\textbf{Embedding Virtual Animals within Camera Trap Imagery.} To finally produce synthetic data that include our virtual individuals and reflect real-world camera trap scenarios, we put forward an image generation pipeline that includes illumination control, background insertion, camera viewpoint variation, and image quality perturbations. These components collectively ensure high fidelity and diversity in the rendered output samples.
For lighting simulation, we employ HDRI captured from real forests as environmental light sources. Sunlight direction and intensity are dynamically adjusted to mimic natural illumination variability.
To replicate the camera angles typical of trap cameras, six virtual cameras are placed around the 3D model, enabling multi-view image capture and improving spatial coverage and sample diversity (see Fig.~\ref{render} (a)). Backgrounds are sourced from a bank of real camera trap images and embedded with the rendered virtual animals using image harmonisation techniques for lighting and colour consistency to ensure seamless integration into the scene.
In addition, we introduce image quality perturbations such as motion blur, noise, and resolution reduction to simulate common challenges in wildlife imaging, simulating animal movement, low light scenarios, and long-distance capture conditions.}

{\textbf{Synthetic Data and Cross-Modal AI Training.} This comprehensive pipeline combines biologically grounded animal modelling with physically simulated environment adjustments to construct a high-fidelity, multi-view, multi-pose image dataset paired with dermatoglyphic text descriptions of the exact virtual animals depicted. This synthetic dataset is cross-modally aligned by design and, together with annotated and text-enhanced real world training data~(see Section~\ref{data}), allows for training AI architectures that map from visual and/or textual descriptors to animal identities~(see Sections~\ref{DL}-\ref{ES} for full details). We next present and evaluate the performance {of our approach} and describe properties of the resulting AI systems when applied to various animal re-identification tasks (see Fig.\ref{network} (c)).}


\subsection{{RESULT A: Animal ID based on Dermatoglyphic Description is Precise}}\label{ResultA}

\begin{figure}[t]
\centering
\includegraphics[width=\textwidth]{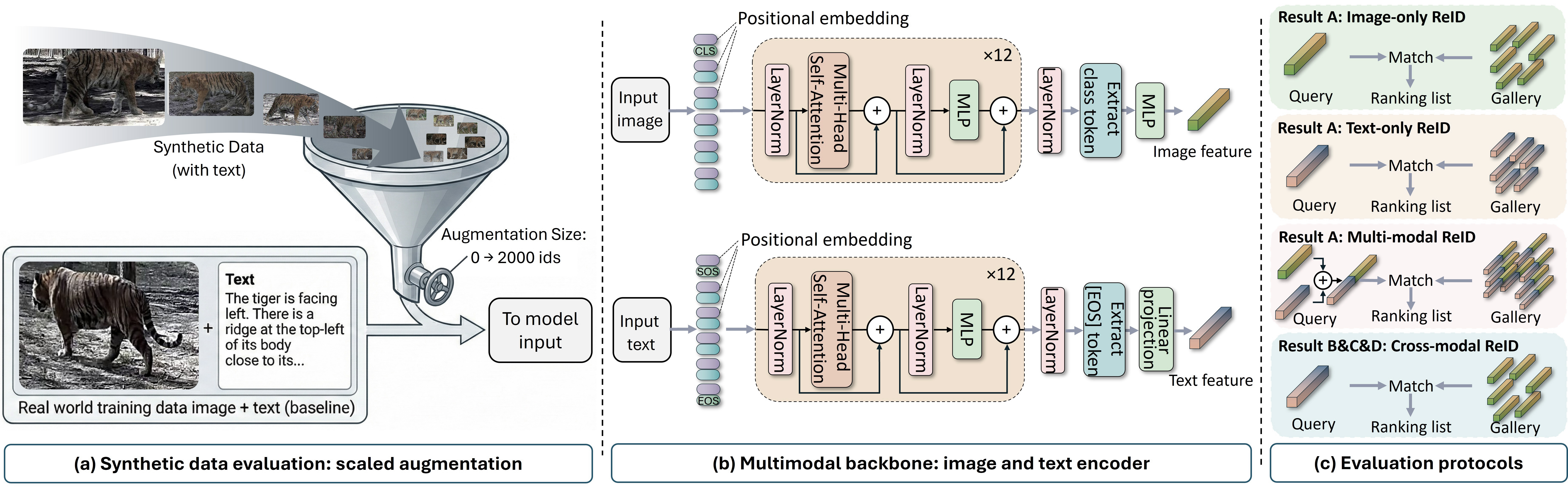}
\caption{{{\textbf{CLIP-based universal re-identification framework and synthetic data evaluation strategy.} 
\textbf{(b)} \textbf{Framework architecture}: The core architecture leverages a CLIP-based dual-stream backbone, employing independent image and text encoders to extract high-dimensional semantic features for diverse downstream tasks.
\textbf{(c)} \textbf{Evaluation protocols}: \textit{Single-modality Re-ID} performs matching within the feature space using a single query modality (text or image) against gallery features of the same modality; \textit{Multi-modal Image-Text Re-ID} utilises concatenated multi-modal features for joint retrieval;  \textit{Cross-modal Re-ID} specifically targets the retrieval of corresponding images using textual feature queries. 
\textbf{(a)} \textbf{Synthetic data evaluation}: To quantify the enhancement efficacy of variable-scale synthetic datasets on real-world scenarios, we established an incremental experimental regime using real data as the baseline. With the test set fixed, synthetic data of varying scales were injected into the training set to systematically evaluate the resulting performance gains in cross-modal Re-ID.}
}}
\label{network}
\end{figure}

{\textbf{Text-only Animal Re-ID.} First, we show that discrete dermatoglyphic text descriptors of ACE encodings derived from visible coat features can enable near-perfect Re-ID of animals --- similar to language-based lab reporting in human forensics. To demonstrate this for tigers, we train a dual-stream architecture based on CLIP ~(see Fig.~\ref{network}~(b)) for the animal re-identification task, in which the text encoder derives textual feature embeddings from the input dermatoglyphic descriptors and conducts similarity matching in the corresponding embedding space.~(see Fig.~\ref{network}~(c)). To evaluate the discriminative capacity of such a text-based minutiae ID system for tigers, we evaluate performance on a real-world animal dataset~(see Section~\ref{data}) combining the ATRW Database~\cite{li2020atrw} and the Camera Trap Database of Tigers from Rajaji
National Park, Uttarakhand, India~\cite{harihar2019tigerrajaji}. The resulting dataset comprises 3,355 images covering 185 real individuals. Each image is manually annotated by us with dermatoglyphic textual descriptor that reflects the exact visual ACE feature sequence visible in each particular image. We report near-perfect performance of $99.8\%$ accuracy on the test set, matching unseen individuals by text descriptors only~(see Section~\ref{data}). We conclude that dermatoglyphic descriptors alone can demonstrably serve as precise, compact, human readable, and image-verifiable discriminators of individual animals.}

\begin{figure}[t]
\centering
\includegraphics[width=\textwidth]{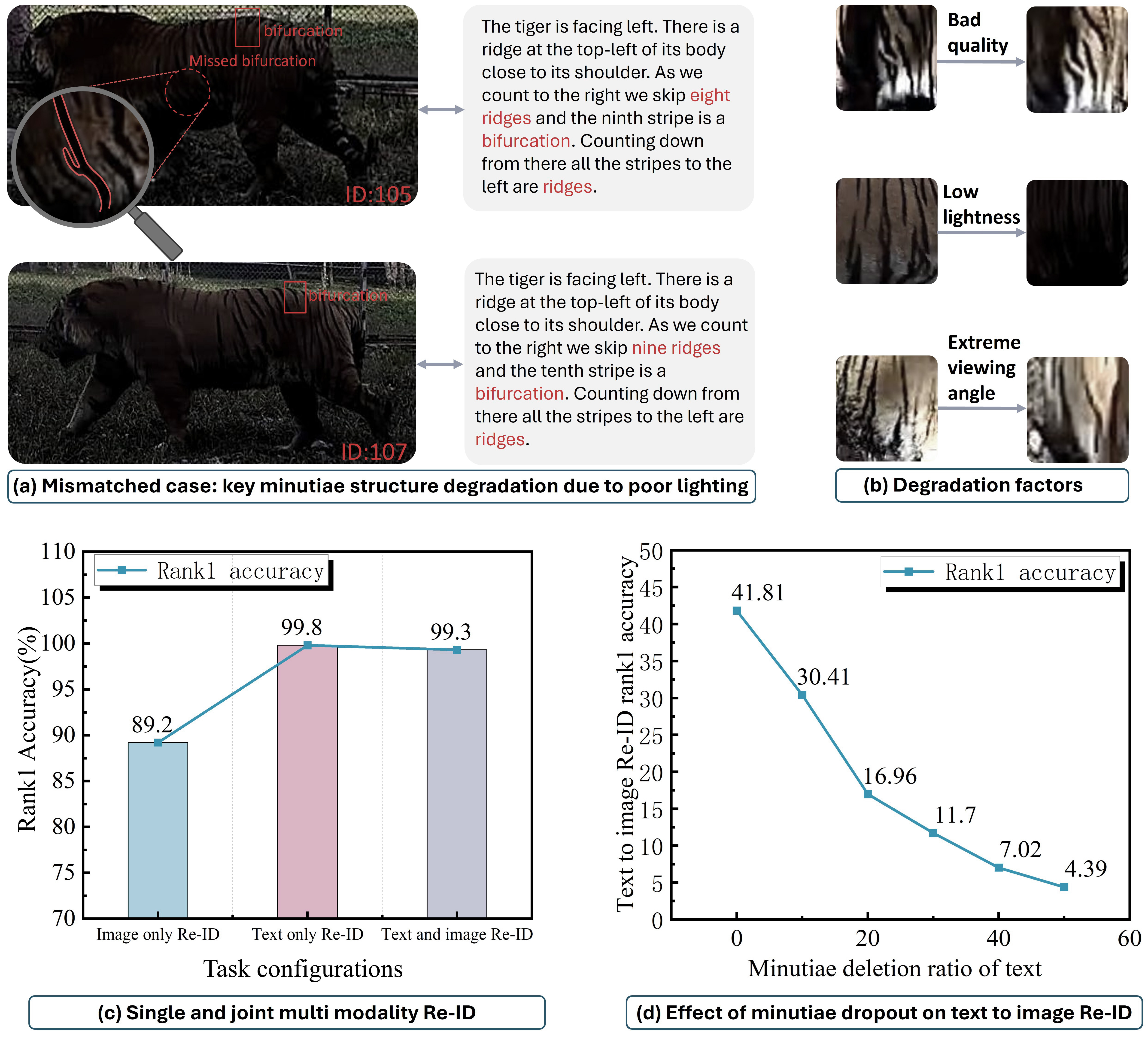}
\caption{{{\textbf{Efficacy of text-assisted Re-ID, failure analysis, and robustness assessment against feature degradation.} 
\textbf{(a) and (b)}: Failure analysis reveals that semantic confusion stems primarily from the degradation of critical minutiae. This phenomenon is largely attributed to environmental factors, including poor imaging quality, insufficient illumination, and extreme viewpoints. \textbf{(c)}: The integration of text assistance significantly enhances identification performance on real-world animal datasets. \textbf{(d)}: We simulated varying degrees of feature loss by randomly culling minutiae from ACE sequences. Results demonstrate an expected progressive decline in cross-modal Re-ID accuracy corresponding to the increasing proportion of minutiae loss.}}}\label{t and i}
\end{figure}

{\textbf{Multi-modal Image-Text Re-ID.} Second, we investigate in how far multi-modal concatenation of dermatoglyphic and visual descriptors can enhance re-identification compared to conventional image-only baselines. 
{The results show that incorporating the text modality substantially boosts visual baseline performance from 89.2\% to 99.3\% and increases sensitivity to subtle inter-individual variation}~(see Fig.~\ref{t and i} (d)). However, the fused image–text representations exhibits an expected, slight performance decline compared to human-annotated discriminative textual features settings {(99.8\%)}. We also note that for visual data collections the cross-modality annotation requirement (human or machine centred) for catalogue and query images remains for this scenario. This is to capture the fine-grained symbolic minutiae information in abstracted textual form at least once per individual.}

{An analysis of rare failure cases reveals that certain distinctive minutiae types—such as bifurcations, convergences, and enclosures—are occasionally degraded into simpler ridge structures due to low image quality. This degradation is primarily caused by noise, poor lighting, or extreme viewing angles, which obscure critical structural details (see Fig.~\ref{t and i} (a) and (b)). As a result, the corresponding textual descriptions lose discriminative content, often converging toward highly similar representations. Collectively, these findings demonstrate that structured textual pattern descriptors not only possess strong stand-alone discriminative power, but also act synergistically with visual inputs to enhance the model’s ability to capture biologically meaningful individual variation.}


\subsection{{RESULT B: Dermatoglyphics enable Novel Text-to-Image Retrieval}}\label{RESULTB}
{\textbf{Animal Image ID from Text Input.} Compared to joint dermatoglyphic text-image Re-ID, automated text-to-image retrieval presents a cross-modal challenge which so far has never been attempted or studied in depth before for animal coats. While the former benefits from the integration of visual and textual modalities, the latter critically relies on direct semantic links between text and corresponding image features across modality boundaries. This setting demands not only fine-grained, minutiae-level alignment between sparse textual descriptions and rich visual features but also the ability to co-learn information that bridges the inherent representation gap between modalities. The setup requires constructing a multi-modal semantic space that co-localises different representations of the same biometric configuration~(Fig. \ref{network}~(c)). Our experimental results~(see Fig.~\ref{t to i}) confirm that such text-to-image retrieval networks can be trained and thereby bridge representational modalities. However, model performance is highly constrained {(21.6\%)} when training on existing, relatively small real-world image collections covering only some hundred tiger IDs. Yet, the introduction of synthetic data based on virtual individuals can reduce the textual-visual domain gap remarkably improving retrieval accuracy.}

\subsection{{RESULT C: Synthetic Training Data enables Practical Cross-modal ID}}\label{RESULT C}

{\textbf{Balancing Real and Synthetic Training Data.} To assess the efficacy of synthetic training data in supporting real-world animal ID tasks, we first establish a performance baseline using only real camera trap data. Subsequently, to explore optimality of real-synthetic data ratios with regard to over-fitting, we progressively augment  real-world training data with increasing amounts of synthetic samples while keeping the test set fixed (see Fig. \ref{network} (a)). With the introduction of synthetic data, retrieval accuracy improves markedly, reaching its peak {(41.8\%)} when approximately 1,000 synthetic identities (around 12,000 text-image pairs) are added to the tiny baseline {(21.6\%)} of 165 identities~(Fig.~\ref{t to i}). This highlights the significant undersampling of animal ID-spaces when only small real-world datasets are used to construct it, but also shows the limits of synthesis where further data generation results in performance degradation over-fitting to the synthetic domain.}

{\textbf{Analysis of Dermatoglyphic Text Descriptor Robustness.} We next assess Re-ID model robustness under degrading ACE descriptors induced by inaccessibility of minutiae information during visual sensor acquisition or human observation. Partial animal occlusions, information loss due to hard shadows or insufficient lighting, or extreme viewing angles are all examples of unrecoverable failure modes~(see Fig.~\ref{t and i}~(b)). We systematically evaluated performance across varying degrees of ACE feature loss to quantify the impact of such information degradation on dermatoglyphic test descriptors. Specifically, we adopted as baseline our best-performing cross-modal retrieval model described above, and simulated degradation by randomly removing minutiae from ACE training sequences at specified proportions. Results reveal a progressive decline {from 41.8\% to only 4.4\%} quite quickly in cross-modal re-identification accuracy with increasing loss of minutiae in the underlying ACE sequence encoded~(see Fig.~\ref{t and i}~(c)). This outcome highlights the model's reliance on sequence information induced by the ACE scheme such that even partial omission substantially impairs precise matching. Quality human observations or recordings from camera traps themselves will therefore remain a core pillar for successful animal identification even under the presented dermatoglyphically enhanced paradigm. Nevertheless, considering potential data uncertainties already during network training can, as we show next, counteract performance degradation to some degree.
}

\subsection{{RESULT D: ACE Anchor-Variations during Training improve Robustness}}\label{RESULT D}
\begin{figure}[t]
\centering
\includegraphics[width=1\textwidth]{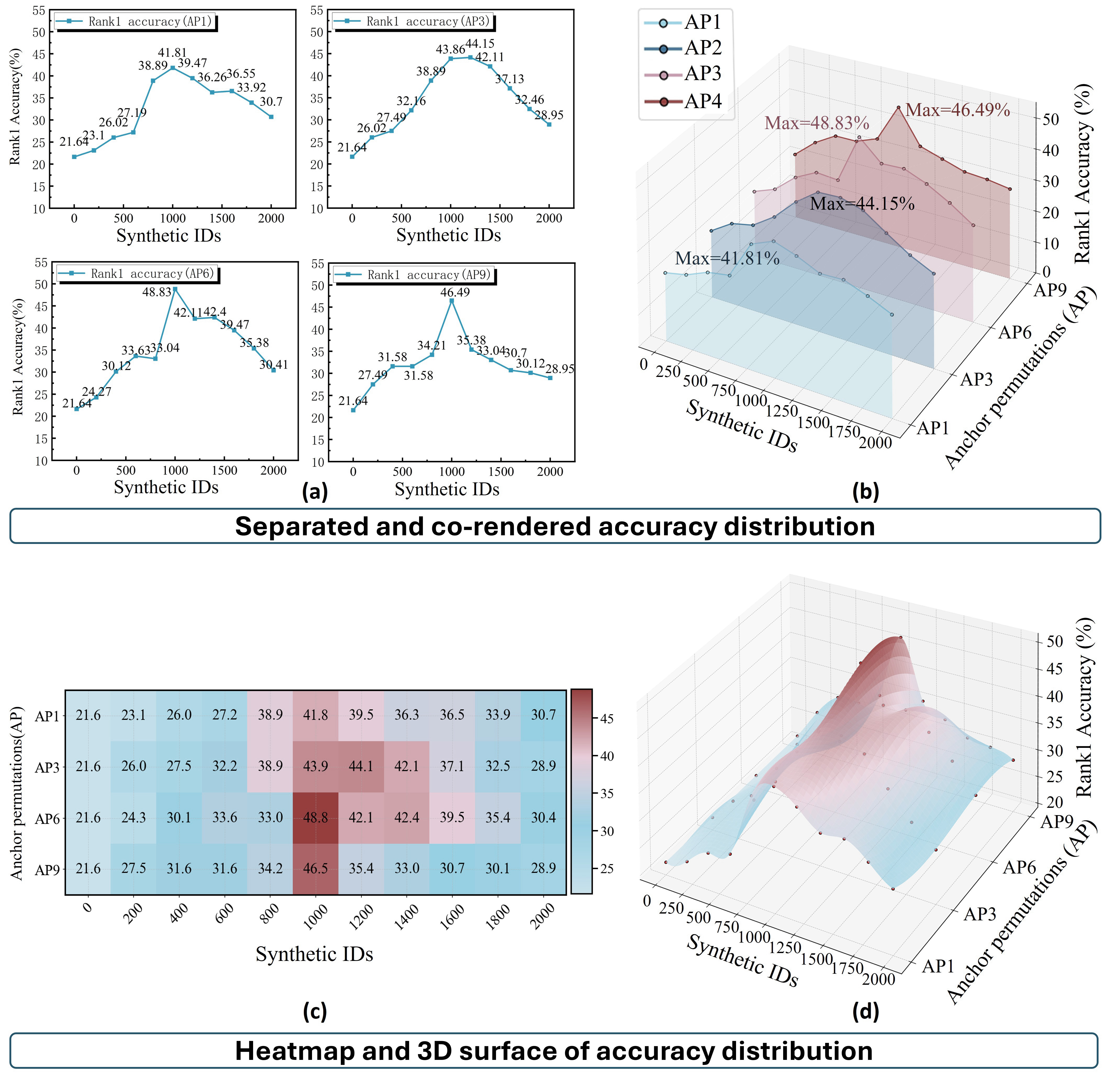}
\caption{{\textbf{Influence of anchor permutation and synthetic data scale on text-to-image Re-ID performance.} 
\textbf{(a)}:  Accuracy trends under different anchor permutation settings as synthetic data scale increases.
\textbf{(b)}: 3D line plot showing joint effects of anchor permutation and data scale on Re-ID accuracy.
\textbf{(c)}: Heatmap illustrating the overall accuracy distribution under varying configurations.
\textbf{(d)}: 3D surface plot visualising performance peaks across the parameter space.
}}\label{t to i}
\end{figure}

{\textbf{Permuting Descriptor Anchors.} Due to the potential presence of multiple adjacent minutiae in regions of ACE anchoring, the exact starting minutia of sequence descriptors cannot be uniquely defined in practice, introducing variations in extracted scanning trajectories. To improve the model’s robustness to these positional deviations, we randomly sample the initial minutiae within a bounded region around the picked anchor. This exposes the uncertainty in path initialisation directly to the training process. Experimental results demonstrate that sampling sequences from about a handful of different starting minutiae~{(AP6)} during training enhances model stability most~({from 41.8\% to 48.8\%,} see Fig. \ref{t to i}). Beyond this point, accuracy declines again.
These findings indicate that, while anchor variation improves tolerance to initialisation noise, consideration beyond a limited local area degrades performance. Bringing both anchor variations and synthetic training data together to enhance network performance results in a system {with 48.8\%} Top-1 retrieval accuracy. This offers a completely new, viable cross-modal coat-pattern identification option sufficient to extend and compliment traditional visual Re-ID pipelines.}

\section{Discussion}\label{sec3}

\subsection{Summary}\label{subsec3.1}
{We introduce a structured dermatoglyphic recognition framework tailored for animal individual identification utilising coat patterns, alongside a controllable text-image co-synthesis pipeline capable of generating life-like virtual individual animals at large-scale in camera trap settings.
Extensive experiments on the sample species~\textit{Panthera tigris} demonstrate the effectiveness of our method across unimodal, multimodal, and for the first time cross-modal minutiae-based animal re-identification task. Furthermore, the approach consistently improves performance and generalisation on real-world datasets, indicating strong potential for practical deployment for the sample species and beyond.}

\subsection{{Explainable Dermatoglyphics as a Bridge between Animal Identity Modalities}}\label{subsec3.2}
{\textbf{Human-readable Codes beyond the Vision-Only Domain.} The presented system and experiments show that recasting coat pattern texture information into dermatoglyphic primitives, namely minutiae and spatial sequences thereof, provides a symbolic representation of animal identities that is modality-independent. Akin to forensic reporting, this work shows that individuality of stripe-textured animals like tigers can be expressed both visually~via sensor imagery as well as symbolically via ACE sequences and reports that describe them in specialist forensic language. Whilst non-computational work in the 1970s by Klingel, Petersen and other ecologists~\cite{petersen1972identification} pioneered this idea of animal coat encoding with symbolic primitives used in capture-mark-recapture~(CMR) studies, our work goes further: we show that this concept can be cast into effective cross-modal AI frameworks and extend current attempts of explainable animal biometrics~\cite{shrack2025pairwise} beyond the purely visual domain. Whilst Klingel's symbolic `stripe codes'~\cite{petersen1972identification} mainly supported the efficient organisation of paper-based population catalogues, the presented approach extends established forensic approaches and automates animal identity retrieval using symbolic dermatoglyphics across modality boundaries. As a result, matching of textual coat descriptions directly to camera trap images is now demonstrably possible.} 

{\textbf{Explainable Re-ID and Realm of Applicability.} The well-recognised universality of dermatoglyphic primitives across Turing-like stripe coats~\cite{burghardt2008visual} may, moreover, offer a more general computational basis for identity data representation in multimodal and cross-modal operation -- exemplified here for tigers. The exact realm of applicability forms a compelling programme for ongoing research. In addition, we note that this work does \textit{not} investigate bidirectional mappings of textual and visual ID domains, such as needed for full dermatoglyphic sensor-to-ID automation without humans-in-the-loop. Our focus, instead, is to \textit{establish} the utility of a fully human-readable and explainable representation for bridging modalities in animal identity retrieval and for enhancing vision-only animal biometrics. Our framework provides explainability of the matching process from language report, to ACE sequence, to minutiae localisation, right down to the captured sensor image of an animal coat. It follows a forensically established pipeline of evidence trailing. This is critical, not only for AI trust building with practitioners in conservation and ecology~\cite{shrack2025pairwise, han2023synergistic}, but for grounding the science of capture-recapture modelling~\cite{kodi2024ghostbusting} so that failure modes in AI identity matching can be investigated and pin-pointed -- both symbolically as well as in the underlying sensor evidence itself.}

\subsection{{Limitations of Generative AI and Necessity for Controllable Dual Synthesis}}\label{subsec3.3}

{\textbf{Inadequacies of Parameter-free Synthesis Options.} 
{The evolution of generative AI systems, particularly diffusion and vision-language models for photorealistic image generation~\cite{esser2024scaling,zhang2023adding,betker2023improving,comanici2025gemini} has been rapid. Yet,} limitations regarding identity-consistent animal coat rendering across images still prohibit their utilisation instead of the classic, proposed co-synthesis pipeline for dermatological sample generation for biometric ID network training. While after significant training efforts identity-consistency for virtual persons can now be achieved relatively reliably~\cite{esser2024scaling}, this does not yet apply to rendered animal coats. Our synthesis framework, in contrast, establishes an explicit mapping between visual structure and dermatoglyphic textual description at minutiae level, building on fine-grained, semantically stable stripe rendering using expert language utilisation and parametric, species-specific 3D animal modelling~\cite{SideFX2023Houdini}.}


{\textbf{Potential for direct Vision-language Animal Synthesis Models.} To {date, even the largest and most sophisticated diffusion models~\cite{esser2024scaling,zhang2023adding,betker2023improving} lack both the mechanisms for precise, multi-image-consistent control over pattern morphology and the scientific terminology to control pattern details inside vision-language co-synthesis. Overall, this prevents} consistent downstream conversion into dual modality descriptors as essentially required for cross-modal system training. Nevertheless, there are no theoretical reasons why such efforts cannot succeed and one would indeed hope for the development of non-parametric, generative AIs capable of identity-aware coat pattern rendering and bi-directional, cross-modal identity representation.}

\bibliography{sn-bibliography}


\begin{thebibliography}{73}
\ifx \bisbn   \undefined \def \bisbn  #1{ISBN #1}\fi
\ifx \binits  \undefined \def \binits#1{#1}\fi
\ifx \bauthor  \undefined \def \bauthor#1{#1}\fi
\ifx \batitle  \undefined \def \batitle#1{#1}\fi
\ifx \bjtitle  \undefined \def \bjtitle#1{#1}\fi
\ifx \bvolume  \undefined \def \bvolume#1{\textbf{#1}}\fi
\ifx \byear  \undefined \def \byear#1{#1}\fi
\ifx \bissue  \undefined \def \bissue#1{#1}\fi
\ifx \bfpage  \undefined \def \bfpage#1{#1}\fi
\ifx \blpage  \undefined \def \blpage #1{#1}\fi
\ifx \burl  \undefined \def \burl#1{\textsf{#1}}\fi
\ifx \doiurl  \undefined \def \doiurl#1{\url{https://doi.org/#1}}\fi
\ifx \betal  \undefined \def \betal{\textit{et al.}}\fi
\ifx \binstitute  \undefined \def \binstitute#1{#1}\fi
\ifx \binstitutionaled  \undefined \def \binstitutionaled#1{#1}\fi
\ifx \bctitle  \undefined \def \bctitle#1{#1}\fi
\ifx \beditor  \undefined \def \beditor#1{#1}\fi
\ifx \bpublisher  \undefined \def \bpublisher#1{#1}\fi
\ifx \bbtitle  \undefined \def \bbtitle#1{#1}\fi
\ifx \bedition  \undefined \def \bedition#1{#1}\fi
\ifx \bseriesno  \undefined \def \bseriesno#1{#1}\fi
\ifx \blocation  \undefined \def \blocation#1{#1}\fi
\ifx \bsertitle  \undefined \def \bsertitle#1{#1}\fi
\ifx \bsnm \undefined \def \bsnm#1{#1}\fi
\ifx \bsuffix \undefined \def \bsuffix#1{#1}\fi
\ifx \bparticle \undefined \def \bparticle#1{#1}\fi
\ifx \barticle \undefined \def \barticle#1{#1}\fi
\bibcommenthead
\ifx \bconfdate \undefined \def \bconfdate #1{#1}\fi
\ifx \botherref \undefined \def \botherref #1{#1}\fi
\ifx \url \undefined \def \url#1{\textsf{#1}}\fi
\ifx \bchapter \undefined \def \bchapter#1{#1}\fi
\ifx \bbook \undefined \def \bbook#1{#1}\fi
\ifx \bcomment \undefined \def \bcomment#1{#1}\fi
\ifx \oauthor \undefined \def \oauthor#1{#1}\fi
\ifx \citeauthoryear \undefined \def \citeauthoryear#1{#1}\fi
\ifx \endbibitem  \undefined \def \endbibitem {}\fi
\ifx \bconflocation  \undefined \def \bconflocation#1{#1}\fi
\ifx \arxivurl  \undefined \def \arxivurl#1{\textsf{#1}}\fi
\csname PreBibitemsHook\endcsname

\bibitem[\protect\citeauthoryear{K{\"u}hl and Burghardt}{2013}]{kuhl2013animal}
\begin{barticle}
\bauthor{\bsnm{K{\"u}hl}, \binits{H.S.}},
\bauthor{\bsnm{Burghardt}, \binits{T.}}:
\batitle{Animal biometrics: quantifying and detecting phenotypic appearance}.
\bjtitle{Trends in Ecology \& Evolution}
\bvolume{28}(\bissue{7}),
\bfpage{432}--\blpage{441}
(\byear{2013})
\end{barticle}
\endbibitem

\bibitem[\protect\citeauthoryear{Chen et~al.}{2021}]{chen2021data}
\begin{bchapter}
\bauthor{\bsnm{Chen}, \binits{W.}},
\bauthor{\bsnm{Zhang}, \binits{W.}},
\bauthor{\bsnm{Liu}, \binits{D.}},
\bauthor{\bsnm{Li}, \binits{W.}},
\bauthor{\bsnm{Shi}, \binits{X.}},
\bauthor{\bsnm{Fang}, \binits{F.}}:
\bctitle{Data-driven multimodal patrol planning for anti-poaching}.
In: \bbtitle{Proceedings of the AAAI Conference on Artificial Intelligence},
pp. \bfpage{15270}--\blpage{15277}
(\byear{2021})
\end{bchapter}
\endbibitem

\bibitem[\protect\citeauthoryear{Tuia et~al.}{2022}]{tuia2022perspectives}
\begin{barticle}
\bauthor{\bsnm{Tuia}, \binits{D.}},
\bauthor{\bsnm{Kellenberger}, \binits{B.}},
\bauthor{\bsnm{Beery}, \binits{S.}},
\bauthor{\bsnm{Costelloe}, \binits{B.R.}},
\bauthor{\bsnm{Zuffi}, \binits{S.}},
\bauthor{\bsnm{Risse}, \binits{B.}},
\bauthor{\bsnm{Mathis}, \binits{A.}},
\bauthor{\bsnm{Mathis}, \binits{M.W.}},
\bauthor{\bsnm{Van~Langevelde}, \binits{F.}},
\bauthor{\bsnm{Burghardt}, \binits{T.}}, \betal:
\batitle{Perspectives in machine learning for wildlife conservation}.
\bjtitle{Nature Communications}
\bvolume{13}(\bissue{1}),
\bfpage{792}
(\byear{2022})
\end{barticle}
\endbibitem

\bibitem[\protect\citeauthoryear{Hebblewhite and Haydon}{2010}]{hebblewhite2010distinguishing}
\begin{barticle}
\bauthor{\bsnm{Hebblewhite}, \binits{M.}},
\bauthor{\bsnm{Haydon}, \binits{D.T.}}:
\batitle{Distinguishing technology from biology: a critical review of the use of gps telemetry data in ecology}.
\bjtitle{Philosophical Transactions of the Royal Society B: Biological Sciences}
\bvolume{365}(\bissue{1550}),
\bfpage{2303}--\blpage{2312}
(\byear{2010})
\end{barticle}
\endbibitem

\bibitem[\protect\citeauthoryear{Walker et~al.}{2011}]{walker2011review}
\begin{barticle}
\bauthor{\bsnm{Walker}, \binits{K.A.}},
\bauthor{\bsnm{Trites}, \binits{A.W.}},
\bauthor{\bsnm{Haulena}, \binits{M.}},
\bauthor{\bsnm{Weary}, \binits{D.M.}}:
\batitle{A review of the effects of different marking and tagging techniques on marine mammals}.
\bjtitle{Wildlife Research}
\bvolume{39}(\bissue{1}),
\bfpage{15}--\blpage{30}
(\byear{2011})
\end{barticle}
\endbibitem

\bibitem[\protect\citeauthoryear{Kays et~al.}{2015}]{kays2015terrestrial}
\begin{barticle}
\bauthor{\bsnm{Kays}, \binits{R.}},
\bauthor{\bsnm{Crofoot}, \binits{M.C.}},
\bauthor{\bsnm{Jetz}, \binits{W.}},
\bauthor{\bsnm{Wikelski}, \binits{M.}}:
\batitle{Terrestrial animal tracking as an eye on life and planet}.
\bjtitle{Science}
\bvolume{348}(\bissue{6240}),
\bfpage{2478}
(\byear{2015})
\end{barticle}
\endbibitem

\bibitem[\protect\citeauthoryear{Beery et~al.}{2019}]{beery2019efficient}
\begin{botherref}
\oauthor{\bsnm{Beery}, \binits{S.}},
\oauthor{\bsnm{Morris}, \binits{D.}},
\oauthor{\bsnm{Yang}, \binits{S.}}:
Efficient pipeline for camera trap image review.
Preprint at https://doi.org/10.48550/arXiv.1907.06772
(2019)
\end{botherref}
\endbibitem

\bibitem[\protect\citeauthoryear{Leorna and Brinkman}{2022}]{leorna2022human}
\begin{barticle}
\bauthor{\bsnm{Leorna}, \binits{S.}},
\bauthor{\bsnm{Brinkman}, \binits{T.}}:
\batitle{Human vs. machine: Detecting wildlife in camera trap images}.
\bjtitle{Ecological Informatics}
\bvolume{72},
\bfpage{101876}
(\byear{2022})
\end{barticle}
\endbibitem

\bibitem[\protect\citeauthoryear{Gadot et~al.}{2024}]{gadot2024crop}
\begin{barticle}
\bauthor{\bsnm{Gadot}, \binits{T.}},
\bauthor{\bsnm{Istrate}, \binits{{\c S}.}},
\bauthor{\bsnm{Kim}, \binits{H.}},
\bauthor{\bsnm{Morris}, \binits{D.}},
\bauthor{\bsnm{Beery}, \binits{S.}},
\bauthor{\bsnm{Birch}, \binits{T.}},
\bauthor{\bsnm{Ahumada}, \binits{J.}}:
\batitle{To crop or not to crop: Comparing whole-image and cropped classification on a large dataset of camera trap images}.
\bjtitle{IET Computer Vision}
\bvolume{18}(\bissue{8}),
\bfpage{1193}--\blpage{1208}
(\byear{2024})
\end{barticle}
\endbibitem

\bibitem[\protect\citeauthoryear{Moskvyak et~al.}{2021}]{moskvyak2021robust}
\begin{bchapter}
\bauthor{\bsnm{Moskvyak}, \binits{O.}},
\bauthor{\bsnm{Maire}, \binits{F.}},
\bauthor{\bsnm{Dayoub}, \binits{F.}},
\bauthor{\bsnm{Armstrong}, \binits{A.O.}},
\bauthor{\bsnm{Baktashmotlagh}, \binits{M.}}:
\bctitle{Robust re-identification of manta rays from natural markings by learning pose invariant embeddings}.
In: \bbtitle{Digital Image Computing: Techniques and Applications},
pp. \bfpage{1}--\blpage{8}
(\byear{2021})
\end{bchapter}
\endbibitem

\bibitem[\protect\citeauthoryear{Schneider et~al.}{2022}]{schneider2022similarity}
\begin{barticle}
\bauthor{\bsnm{Schneider}, \binits{S.}},
\bauthor{\bsnm{Taylor}, \binits{G.W.}},
\bauthor{\bsnm{Kremer}, \binits{S.C.}}:
\batitle{Similarity learning networks for animal individual re-identification: an ecological perspective}.
\bjtitle{Mammalian Biology}
\bvolume{102}(\bissue{3}),
\bfpage{899}--\blpage{914}
(\byear{2022})
\end{barticle}
\endbibitem

\bibitem[\protect\citeauthoryear{Reynolds et~al.}{2025}]{reynolds2025potential}
\begin{barticle}
\bauthor{\bsnm{Reynolds}, \binits{S.A.}},
\bauthor{\bsnm{Beery}, \binits{S.}},
\bauthor{\bsnm{Burgess}, \binits{N.}},
\bauthor{\bsnm{Burgman}, \binits{M.}},
\bauthor{\bsnm{Butchart}, \binits{S.H.}},
\bauthor{\bsnm{Cooke}, \binits{S.J.}},
\bauthor{\bsnm{Coomes}, \binits{D.}},
\bauthor{\bsnm{Danielsen}, \binits{F.}},
\bauthor{\bsnm{Di~Minin}, \binits{E.}},
\bauthor{\bsnm{Dur{\'a}n}, \binits{A.P.}}, \betal:
\batitle{The potential for ai to revolutionize conservation: a horizon scan}.
\bjtitle{Trends in ecology \& evolution}
\bvolume{40}(\bissue{2}),
\bfpage{191}--\blpage{207}
(\byear{2025})
\end{barticle}
\endbibitem

\bibitem[\protect\citeauthoryear{Kondo and Miura}{2010}]{kondo2010reaction}
\begin{barticle}
\bauthor{\bsnm{Kondo}, \binits{S.}},
\bauthor{\bsnm{Miura}, \binits{T.}}:
\batitle{Reaction-diffusion model as a framework for understanding biological pattern formation}.
\bjtitle{Science}
\bvolume{329}(\bissue{5999}),
\bfpage{1616}--\blpage{1620}
(\byear{2010})
\end{barticle}
\endbibitem

\bibitem[\protect\citeauthoryear{Eizirik et~al.}{2010}]{eizirik2010defining}
\begin{barticle}
\bauthor{\bsnm{Eizirik}, \binits{E.}},
\bauthor{\bsnm{David}, \binits{V.A.}},
\bauthor{\bsnm{Buckley-Beason}, \binits{V.}},
\bauthor{\bsnm{Roelke}, \binits{M.E.}},
\bauthor{\bsnm{Schaffer}, \binits{A.A.}},
\bauthor{\bsnm{Hannah}, \binits{S.S.}},
\bauthor{\bsnm{Narfstrom}, \binits{K.}},
\bauthor{\bsnm{O'Brien}, \binits{S.J.}},
\bauthor{\bsnm{Menotti-Raymond}, \binits{M.}}:
\batitle{Defining and mapping mammalian coat pattern genes: multiple genomic regions implicated in domestic cat stripes and spots}.
\bjtitle{Genetics}
\bvolume{184}(\bissue{1}),
\bfpage{267}--\blpage{275}
(\byear{2010})
\end{barticle}
\endbibitem

\bibitem[\protect\citeauthoryear{Kondo}{2002}]{kondo2002reaction}
\begin{barticle}
\bauthor{\bsnm{Kondo}, \binits{S.}}:
\batitle{The reaction-diffusion system: A mechanism for autonomous pattern formation in the animal skin}.
\bjtitle{Genes to Cells}
\bvolume{7}(\bissue{6}),
\bfpage{535}--\blpage{541}
(\byear{2002})
\end{barticle}
\endbibitem

\bibitem[\protect\citeauthoryear{Burghardt}{2008}]{burghardt2008visual}
\begin{botherref}
\oauthor{\bsnm{Burghardt}, \binits{T.}}:
Visual animal biometrics.
Ph.d. thesis,
University of Bristol,
Bristol, UK
(2008)
\end{botherref}
\endbibitem

\bibitem[\protect\citeauthoryear{Bolger et~al.}{2012}]{bolger2012computer}
\begin{barticle}
\bauthor{\bsnm{Bolger}, \binits{D.T.}},
\bauthor{\bsnm{Morrison}, \binits{T.A.}},
\bauthor{\bsnm{Vance}, \binits{B.}},
\bauthor{\bsnm{Lee}, \binits{D.}},
\bauthor{\bsnm{Farid}, \binits{H.}}:
\batitle{A computer-assisted system for photographic mark--recapture analysis}.
\bjtitle{Methods in Ecology and Evolution}
\bvolume{3}(\bissue{5}),
\bfpage{813}--\blpage{822}
(\byear{2012})
\end{barticle}
\endbibitem

\bibitem[\protect\citeauthoryear{Cihan et~al.}{2023}]{cihan2023identification}
\begin{barticle}
\bauthor{\bsnm{Cihan}, \binits{P.}},
\bauthor{\bsnm{Saygili}, \binits{A.}},
\bauthor{\bsnm{Ozmen}, \binits{N.E.}},
\bauthor{\bsnm{Akyuzlu}, \binits{M.}}:
\batitle{Identification and recognition of animals from biometric markers using computer vision approaches: A review}.
\bjtitle{Kafkas Univ Vet Fak Derg}
\bvolume{29}(\bissue{6}),
\bfpage{581}--\blpage{593}
(\byear{2023})
\end{barticle}
\endbibitem

\bibitem[\protect\citeauthoryear{Jain}{2005}]{jain2005biometric}
\begin{bchapter}
\bauthor{\bsnm{Jain}, \binits{A.K.}}:
\bctitle{Biometric recognition: how do i know who you are?}
In: \bbtitle{Proceedings of the International Conference on Image Analysis and Processing},
pp. \bfpage{19}--\blpage{26}
(\byear{2005})
\end{bchapter}
\endbibitem

\bibitem[\protect\citeauthoryear{Burton et~al.}{2015}]{burton2015wildlife}
\begin{barticle}
\bauthor{\bsnm{Burton}, \binits{A.C.}},
\bauthor{\bsnm{Neilson}, \binits{E.}},
\bauthor{\bsnm{Moreira}, \binits{D.}},
\bauthor{\bsnm{Ladle}, \binits{A.}},
\bauthor{\bsnm{Steenweg}, \binits{R.}},
\bauthor{\bsnm{Fisher}, \binits{J.T.}},
\bauthor{\bsnm{Bayne}, \binits{E.}},
\bauthor{\bsnm{Boutin}, \binits{S.}}:
\batitle{Wildlife camera trapping: a review and recommendations for linking surveys to ecological processes}.
\bjtitle{Journal of Applied Ecology}
\bvolume{52}(\bissue{3}),
\bfpage{675}--\blpage{685}
(\byear{2015})
\end{barticle}
\endbibitem

\bibitem[\protect\citeauthoryear{Ravoor and Sudarshan}{2020}]{ravoor2020deep}
\begin{barticle}
\bauthor{\bsnm{Ravoor}, \binits{P.C.}},
\bauthor{\bsnm{Sudarshan}, \binits{T.}}:
\batitle{Deep learning methods for multi-species animal re-identification and tracking--a survey}.
\bjtitle{Computer Science Review}
\bvolume{38},
\bfpage{100289}
(\byear{2020})
\end{barticle}
\endbibitem

\bibitem[\protect\citeauthoryear{Ma et~al.}{2025}]{ma2025deep}
\begin{barticle}
\bauthor{\bsnm{Ma}, \binits{Y.}},
\bauthor{\bsnm{Tan}, \binits{M.}},
\bauthor{\bsnm{Liu}, \binits{X.}},
\bauthor{\bsnm{Zhang}, \binits{Y.}},
\bauthor{\bsnm{Xu}, \binits{Z.}},
\bauthor{\bsnm{Sun}, \binits{W.}},
\bauthor{\bsnm{Ge}, \binits{J.}},
\bauthor{\bsnm{Feng}, \binits{L.}}:
\batitle{Deep learning for amur tiger re-identification in camera traps: A tool assisting population monitoring and spatio-temporal analysis}.
\bjtitle{Ecological Indicators}
\bvolume{171},
\bfpage{113227}
(\byear{2025})
\end{barticle}
\endbibitem

\bibitem[\protect\citeauthoryear{Li et~al.}{2020}]{li2020atrw}
\begin{bchapter}
\bauthor{\bsnm{Li}, \binits{S.}},
\bauthor{\bsnm{Li}, \binits{J.}},
\bauthor{\bsnm{Tang}, \binits{H.}},
\bauthor{\bsnm{Qian}, \binits{R.}},
\bauthor{\bsnm{Lin}, \binits{W.}}:
\bctitle{Atrw: A benchmark for amur tiger re-identification in the wild}.
In: \bbtitle{Proceedings of the ACM International Conference on Multimedia},
pp. \bfpage{2590}--\blpage{2598}
(\byear{2020})
\end{bchapter}
\endbibitem

\bibitem[\protect\citeauthoryear{Bedetti et~al.}{2020}]{bedetti2020system}
\begin{barticle}
\bauthor{\bsnm{Bedetti}, \binits{A.}},
\bauthor{\bsnm{Greyling}, \binits{C.}},
\bauthor{\bsnm{Paul}, \binits{B.}},
\bauthor{\bsnm{Blondeau}, \binits{J.}},
\bauthor{\bsnm{Clark}, \binits{A.}},
\bauthor{\bsnm{Malin}, \binits{H.}},
\bauthor{\bsnm{Horne}, \binits{J.}},
\bauthor{\bsnm{Makukule}, \binits{R.}},
\bauthor{\bsnm{Wilmot}, \binits{J.}},
\bauthor{\bsnm{Eggeling}, \binits{T.}}, \betal:
\batitle{System for elephant ear-pattern knowledge (seek) to identify individual african elephants}.
\bjtitle{Pachyderm}
\bvolume{61},
\bfpage{63}--\blpage{77}
(\byear{2020})
\end{barticle}
\endbibitem

\bibitem[\protect\citeauthoryear{Prinsloo et~al.}{2021}]{prinsloo2021unique}
\begin{barticle}
\bauthor{\bsnm{Prinsloo}, \binits{N.D.}},
\bauthor{\bsnm{Postma}, \binits{M.}},
\bauthor{\bsnm{Bruyn}, \binits{P.N.}}:
\batitle{How unique is unique? quantifying geometric differences in stripe patterns of cape mountain zebra, equus zebra zebra (perissodactyla: Equidae)}.
\bjtitle{Zoological Journal of the Linnean Society}
\bvolume{191}(\bissue{2}),
\bfpage{612}--\blpage{625}
(\byear{2021})
\end{barticle}
\endbibitem

\bibitem[\protect\citeauthoryear{Petso et~al.}{2022}]{petso2022review}
\begin{barticle}
\bauthor{\bsnm{Petso}, \binits{T.}},
\bauthor{\bsnm{Jamisola~Jr}, \binits{R.S.}},
\bauthor{\bsnm{Mpoeleng}, \binits{D.}}:
\batitle{Review on methods used for wildlife species and individual identification}.
\bjtitle{European Journal of Wildlife Research}
\bvolume{68}(\bissue{1}),
\bfpage{3}
(\byear{2022})
\end{barticle}
\endbibitem

\bibitem[\protect\citeauthoryear{Cheeseman et~al.}{2022}]{cheeseman2022advanced}
\begin{barticle}
\bauthor{\bsnm{Cheeseman}, \binits{T.}},
\bauthor{\bsnm{Southerland}, \binits{K.}},
\bauthor{\bsnm{Park}, \binits{J.}},
\bauthor{\bsnm{Olio}, \binits{M.}},
\bauthor{\bsnm{Flynn}, \binits{K.}},
\bauthor{\bsnm{Calambokidis}, \binits{J.}},
\bauthor{\bsnm{Jones}, \binits{L.}},
\bauthor{\bsnm{Garrigue}, \binits{C.}},
\bauthor{\bsnm{Frisch~Jordan}, \binits{A.}},
\bauthor{\bsnm{Howard}, \binits{A.}}, \betal:
\batitle{Advanced image recognition: a fully automated, high-accuracy photo-identification matching system for humpback whales}.
\bjtitle{Mammalian Biology}
\bvolume{102}(\bissue{3}),
\bfpage{915}--\blpage{929}
(\byear{2022})
\end{barticle}
\endbibitem

\bibitem[\protect\citeauthoryear{G{\'o}mez-Vargas et~al.}{2023}]{gomez2023re}
\begin{barticle}
\bauthor{\bsnm{G{\'o}mez-Vargas}, \binits{N.}},
\bauthor{\bsnm{Alonso-Fern{\'a}ndez}, \binits{A.}},
\bauthor{\bsnm{Blanquero}, \binits{R.}},
\bauthor{\bsnm{Antelo}, \binits{L.T.}}:
\batitle{Re-identification of fish individuals of undulate skate via deep learning within a few-shot context}.
\bjtitle{Ecological Informatics}
\bvolume{75},
\bfpage{102036}
(\byear{2023})
\end{barticle}
\endbibitem

\bibitem[\protect\citeauthoryear{Meguro and Hasumura}{2024}]{meguro2024stripe}
\begin{barticle}
\bauthor{\bsnm{Meguro}, \binits{S.}},
\bauthor{\bsnm{Hasumura}, \binits{T.}}:
\batitle{Stripe pattern differences can be used to distinguish individual adult zebrafish}.
\bjtitle{Plos One}
\bvolume{19}(\bissue{10}),
\bfpage{0311372}
(\byear{2024})
\end{barticle}
\endbibitem

\bibitem[\protect\citeauthoryear{Nepovinnykh et~al.}{2024}]{nepovinnykh2024species}
\begin{barticle}
\bauthor{\bsnm{Nepovinnykh}, \binits{E.}},
\bauthor{\bsnm{Chelak}, \binits{I.}},
\bauthor{\bsnm{Eerola}, \binits{T.}},
\bauthor{\bsnm{Immonen}, \binits{V.}},
\bauthor{\bsnm{K{\"a}lvi{\"a}inen}, \binits{H.}},
\bauthor{\bsnm{Kholiavchenko}, \binits{M.}},
\bauthor{\bsnm{Stewart}, \binits{C.V.}}:
\batitle{Species-agnostic patterned animal re-identification by aggregating deep local features}.
\bjtitle{International Journal of Computer Vision}
\bvolume{132}(\bissue{9}),
\bfpage{4003}--\blpage{4018}
(\byear{2024})
\end{barticle}
\endbibitem

\bibitem[\protect\citeauthoryear{Gholami et~al.}{2025}]{gholami2025ai}
\begin{botherref}
\oauthor{\bsnm{Gholami}, \binits{S.}},
\oauthor{\bsnm{Lavista~Ferres}, \binits{J.}},
\oauthor{\bsnm{Lee}, \binits{D.}}:
AI-powered Software for Identifying Individual Giraffes.
Software
(2025).
\url{https://github.com/microsoft/giraffe-identification-ai-tool}
\end{botherref}
\endbibitem

\bibitem[\protect\citeauthoryear{Wang et~al.}{2021}]{wang2021giant}
\begin{barticle}
\bauthor{\bsnm{Wang}, \binits{L.}},
\bauthor{\bsnm{Ding}, \binits{R.}},
\bauthor{\bsnm{Zhai}, \binits{Y.}},
\bauthor{\bsnm{Zhang}, \binits{Q.}},
\bauthor{\bsnm{Tang}, \binits{W.}},
\bauthor{\bsnm{Zheng}, \binits{N.}},
\bauthor{\bsnm{Hua}, \binits{G.}}:
\batitle{Giant panda identification}.
\bjtitle{IEEE Transactions on Image Processing}
\bvolume{30},
\bfpage{2837}--\blpage{2849}
(\byear{2021})
\end{barticle}
\endbibitem

\bibitem[\protect\citeauthoryear{Zheng et~al.}{2022}]{zheng2022wild}
\begin{barticle}
\bauthor{\bsnm{Zheng}, \binits{Z.}},
\bauthor{\bsnm{Zhao}, \binits{Y.}},
\bauthor{\bsnm{Li}, \binits{A.}},
\bauthor{\bsnm{Yu}, \binits{Q.}}:
\batitle{Wild terrestrial animal re-identification based on an improved locally aware transformer with a cross-attention mechanism}.
\bjtitle{Animals}
\bvolume{12}(\bissue{24}),
\bfpage{3503}
(\byear{2022})
\end{barticle}
\endbibitem

\bibitem[\protect\citeauthoryear{Li et~al.}{2024}]{li2024adaptive}
\begin{bchapter}
\bauthor{\bsnm{Li}, \binits{C.}},
\bauthor{\bsnm{Chen}, \binits{S.}},
\bauthor{\bsnm{Ye}, \binits{M.}}:
\bctitle{Adaptive high-frequency transformer for diverse wildlife re-identification}.
In: \bbtitle{Proceedings of the European Conference on Computer Vision},
pp. \bfpage{296}--\blpage{313}
(\byear{2024})
\end{bchapter}
\endbibitem

\bibitem[\protect\citeauthoryear{Han et~al.}{2024}]{han2024multi}
\begin{barticle}
\bauthor{\bsnm{Han}, \binits{Y.}},
\bauthor{\bsnm{Chen}, \binits{K.}},
\bauthor{\bsnm{Wang}, \binits{Y.}},
\bauthor{\bsnm{Liu}, \binits{W.}},
\bauthor{\bsnm{Wang}, \binits{Z.}},
\bauthor{\bsnm{Wang}, \binits{X.}},
\bauthor{\bsnm{Han}, \binits{C.}},
\bauthor{\bsnm{Liao}, \binits{J.}},
\bauthor{\bsnm{Huang}, \binits{K.}},
\bauthor{\bsnm{Cai}, \binits{S.}}, \betal:
\batitle{Multi-animal 3d social pose estimation, identification and behaviour embedding with a few-shot learning framework}.
\bjtitle{Nature Machine Intelligence}
\bvolume{6}(\bissue{1}),
\bfpage{48}--\blpage{61}
(\byear{2024})
\end{barticle}
\endbibitem

\bibitem[\protect\citeauthoryear{Mulero-P{\'a}zm{\'a}ny et~al.}{2025}]{mulero2025addressing}
\begin{barticle}
\bauthor{\bsnm{Mulero-P{\'a}zm{\'a}ny}, \binits{M.}},
\bauthor{\bsnm{Hurtado}, \binits{S.}},
\bauthor{\bsnm{Barba-Gonz{\'a}lez}, \binits{C.}},
\bauthor{\bsnm{Antequera-G{\'o}mez}, \binits{M.L.}},
\bauthor{\bsnm{D{\'\i}az-Ruiz}, \binits{F.}},
\bauthor{\bsnm{Real}, \binits{R.}},
\bauthor{\bsnm{Navas-Delgado}, \binits{I.}},
\bauthor{\bsnm{Aldana-Montes}, \binits{J.F.}}:
\batitle{Addressing significant challenges for animal detection in camera trap images: a novel deep learning-based approach}.
\bjtitle{Scientific Reports}
\bvolume{15}(\bissue{1}),
\bfpage{16191}
(\byear{2025})
\end{barticle}
\endbibitem

\bibitem[\protect\citeauthoryear{Stevenage and Pitfield}{2016}]{stevenage2016fact}
\begin{barticle}
\bauthor{\bsnm{Stevenage}, \binits{S.V.}},
\bauthor{\bsnm{Pitfield}, \binits{C.}}:
\batitle{Fact or friction: Examination of the transparency, reliability and sufficiency of the ace-v method of fingerprint analysis}.
\bjtitle{Forensic Science International}
\bvolume{267},
\bfpage{145}--\blpage{156}
(\byear{2016})
\end{barticle}
\endbibitem

\bibitem[\protect\citeauthoryear{Chauhan}{2017}]{chauhan2017latent}
\begin{barticle}
\bauthor{\bsnm{Chauhan}, \binits{A.}}:
\batitle{Latent palm prints—an appraisal of concealed individualize evidence and its aspect in forensic investigation}.
\bjtitle{Journal of Forensic Sciences \& Criminal Investigation}
\bvolume{6},
\bfpage{555696}
(\byear{2017})
\end{barticle}
\endbibitem

\bibitem[\protect\citeauthoryear{Dhaneshwar et~al.}{2021}]{dhaneshwar2021investigation}
\begin{barticle}
\bauthor{\bsnm{Dhaneshwar}, \binits{R.}},
\bauthor{\bsnm{Kaur}, \binits{M.}},
\bauthor{\bsnm{Kaur}, \binits{M.}}:
\batitle{An investigation of latent fingerprinting techniques}.
\bjtitle{Egyptian Journal of Forensic Sciences}
\bvolume{11}(\bissue{1}),
\bfpage{33}
(\byear{2021})
\end{barticle}
\endbibitem

\bibitem[\protect\citeauthoryear{Needham et~al.}{2022}]{needham2022collaborative}
\begin{barticle}
\bauthor{\bsnm{Needham}, \binits{M.}},
\bauthor{\bsnm{Fieldhouse}, \binits{S.}},
\bauthor{\bsnm{Morris}, \binits{W.}},
\bauthor{\bsnm{Wheeler}, \binits{J.}},
\bauthor{\bsnm{Nicholls}, \binits{G.}}:
\batitle{Collaborative practise in forensic science and academia: The development of a documentation strategy for fingerprint examinations in an english fingerprint bureau in the iso 17025 era}.
\bjtitle{Science \& Justice}
\bvolume{62}(\bissue{3}),
\bfpage{336}--\blpage{348}
(\byear{2022})
\end{barticle}
\endbibitem

\bibitem[\protect\citeauthoryear{Khodadoust et~al.}{2024}]{khodadoust2024enhancing}
\begin{barticle}
\bauthor{\bsnm{Khodadoust}, \binits{J.}},
\bauthor{\bsnm{Monroy}, \binits{R.}},
\bauthor{\bsnm{Medina-P{\'e}rez}, \binits{M.A.}},
\bauthor{\bsnm{Loyola-Gonz{\'a}lez}, \binits{O.}},
\bauthor{\bsnm{Areekul}, \binits{V.}},
\bauthor{\bsnm{Kusakunniran}, \binits{W.}}:
\batitle{Enhancing latent palmprints using frequency domain analysis}.
\bjtitle{Intelligent Systems with Applications}
\bvolume{23},
\bfpage{200414}
(\byear{2024})
\end{barticle}
\endbibitem

\bibitem[\protect\citeauthoryear{Petersen}{1972}]{petersen1972identification}
\begin{barticle}
\bauthor{\bsnm{Petersen}, \binits{J.B.}}:
\batitle{An identification system for zebra (equus burchelli, gray)}.
\bjtitle{African Journal of Ecology}
\bvolume{10}(\bissue{1}),
\bfpage{59}--\blpage{63}
(\byear{1972})
\end{barticle}
\endbibitem

\bibitem[\protect\citeauthoryear{Allen et~al.}{2011}]{allen2011leopard}
\begin{barticle}
\bauthor{\bsnm{Allen}, \binits{W.L.}},
\bauthor{\bsnm{Cuthill}, \binits{I.C.}},
\bauthor{\bsnm{Scott-Samuel}, \binits{N.E.}},
\bauthor{\bsnm{Baddeley}, \binits{R.}}:
\batitle{Why the leopard got its spots: relating pattern development to ecology in felids}.
\bjtitle{Proceedings of the Royal Society B: Biological Sciences}
\bvolume{278}(\bissue{1710}),
\bfpage{1373}--\blpage{1380}
(\byear{2011})
\end{barticle}
\endbibitem

\bibitem[\protect\citeauthoryear{Lee et~al.}{2018}]{lee2018seeing}
\begin{barticle}
\bauthor{\bsnm{Lee}, \binits{D.E.}},
\bauthor{\bsnm{Cavener}, \binits{D.R.}},
\bauthor{\bsnm{Bond}, \binits{M.L.}}:
\batitle{Seeing spots: quantifying mother-offspring similarity and assessing fitness consequences of coat pattern traits in a wild population of giraffes (giraffa camelopardalis)}.
\bjtitle{PeerJ}
\bvolume{6},
\bfpage{5690}
(\byear{2018})
\end{barticle}
\endbibitem

\bibitem[\protect\citeauthoryear{Chan et~al.}{2019}]{chan2019pat}
\begin{barticle}
\bauthor{\bsnm{Chan}, \binits{I.Z.}},
\bauthor{\bsnm{Stevens}, \binits{M.}},
\bauthor{\bsnm{Todd}, \binits{P.A.}}:
\batitle{Pat-geom: a software package for the analysis of animal patterns}.
\bjtitle{Methods in Ecology and Evolution}
\bvolume{10}(\bissue{4}),
\bfpage{591}--\blpage{600}
(\byear{2019})
\end{barticle}
\endbibitem

\bibitem[\protect\citeauthoryear{Staps et~al.}{2023}]{staps2023development}
\begin{barticle}
\bauthor{\bsnm{Staps}, \binits{M.}},
\bauthor{\bsnm{Miller}, \binits{P.W.}},
\bauthor{\bsnm{Tarnita}, \binits{C.E.}},
\bauthor{\bsnm{Mallarino}, \binits{R.}}:
\batitle{Development shapes the evolutionary diversification of rodent stripe patterns}.
\bjtitle{Proceedings of the National Academy of Sciences}
\bvolume{120}(\bissue{45}),
\bfpage{2312077120}
(\byear{2023})
\end{barticle}
\endbibitem

\bibitem[\protect\citeauthoryear{Yager and Amin}{2004}]{yager2004fingerprint}
\begin{barticle}
\bauthor{\bsnm{Yager}, \binits{N.}},
\bauthor{\bsnm{Amin}, \binits{A.}}:
\batitle{Fingerprint verification based on minutiae features: a review}.
\bjtitle{Pattern Analysis and Applications}
\bvolume{7}(\bissue{1}),
\bfpage{94}--\blpage{113}
(\byear{2004})
\end{barticle}
\endbibitem

\bibitem[\protect\citeauthoryear{Martins et~al.}{2024}]{martins2024fingerprint}
\begin{barticle}
\bauthor{\bsnm{Martins}, \binits{N.}},
\bauthor{\bsnm{Silva}, \binits{J.S.}},
\bauthor{\bsnm{Bernardino}, \binits{A.}}:
\batitle{Fingerprint recognition in forensic scenarios}.
\bjtitle{Sensors}
\bvolume{24}(\bissue{2}),
\bfpage{664}
(\byear{2024})
\end{barticle}
\endbibitem

\bibitem[\protect\citeauthoryear{Stoney and Thornton}{1986}]{stoney1986critical}
\begin{barticle}
\bauthor{\bsnm{Stoney}, \binits{D.A.}},
\bauthor{\bsnm{Thornton}, \binits{J.I.}}:
\batitle{A critical analysis of quantitative fingerprint individuality models}.
\bjtitle{Journal of Forensic Sciences}
\bvolume{31}(\bissue{4}),
\bfpage{1187}--\blpage{1216}
(\byear{1986})
\end{barticle}
\endbibitem

\bibitem[\protect\citeauthoryear{Ouyang and Swinney}{1991}]{ouyang1991transition}
\begin{barticle}
\bauthor{\bsnm{Ouyang}, \binits{Q.}},
\bauthor{\bsnm{Swinney}, \binits{H.L.}}:
\batitle{Transition from a uniform state to hexagonal and striped turing patterns}.
\bjtitle{Nature}
\bvolume{352}(\bissue{6336}),
\bfpage{610}--\blpage{612}
(\byear{1991})
\end{barticle}
\endbibitem

\bibitem[\protect\citeauthoryear{Malhotra et~al.}{2020}]{malhotra2020understanding}
\begin{barticle}
\bauthor{\bsnm{Malhotra}, \binits{A.}},
\bauthor{\bsnm{Sankaran}, \binits{A.}},
\bauthor{\bsnm{Vatsa}, \binits{M.}},
\bauthor{\bsnm{Singh}, \binits{R.}},
\bauthor{\bsnm{Morris}, \binits{K.B.}},
\bauthor{\bsnm{Noore}, \binits{A.}}:
\batitle{Understanding ace-v latent fingerprint examination process via eye-gaze analysis}.
\bjtitle{IEEE Transactions on Biometrics, Behavior, and Identity Science}
\bvolume{3}(\bissue{1}),
\bfpage{44}--\blpage{58}
(\byear{2020})
\end{barticle}
\endbibitem

\bibitem[\protect\citeauthoryear{Deshpande and Malemath}{2022}]{deshpande2022study}
\begin{barticle}
\bauthor{\bsnm{Deshpande}, \binits{U.U.}},
\bauthor{\bsnm{Malemath}, \binits{V.}}:
\batitle{A study on automatic latent fingerprint identification system}.
\bjtitle{Journal of Computer Science Research}
\bvolume{4}(\bissue{1}),
\bfpage{38}--\blpage{50}
(\byear{2022})
\end{barticle}
\endbibitem

\bibitem[\protect\citeauthoryear{Shrack et~al.}{2025}]{shrack2025pairwise}
\begin{botherref}
\oauthor{\bsnm{Shrack}, \binits{L.}},
\oauthor{\bsnm{Haucke}, \binits{T.}},
\oauthor{\bsnm{Sala{\"u}n}, \binits{A.}},
\oauthor{\bsnm{Subramonian}, \binits{A.}},
\oauthor{\bsnm{Beery}, \binits{S.}}:
Pairwise Matching of Intermediate Representations for Fine-grained Explainability.
Preprint at https://doi.org/10.48550/arXiv.2503.22881
(2025)
\end{botherref}
\endbibitem

\bibitem[\protect\citeauthoryear{Flavell}{2010}]{flavell2010uv}
\begin{bchapter}
\bauthor{\bsnm{Flavell}, \binits{L.}}:
\bctitle{Uv mapping}.
In: \bbtitle{Beginning Blender: Open Source 3D Modeling, Animation, and Game Design},
pp. \bfpage{97}--\blpage{122}.
\bpublisher{Apress},
\blocation{Berkeley, CA}
(\byear{2010})
\end{bchapter}
\endbibitem

\bibitem[\protect\citeauthoryear{Poranne et~al.}{2017}]{poranne2017autocuts}
\begin{barticle}
\bauthor{\bsnm{Poranne}, \binits{R.}},
\bauthor{\bsnm{Tarini}, \binits{M.}},
\bauthor{\bsnm{Huber}, \binits{S.}},
\bauthor{\bsnm{Panozzo}, \binits{D.}},
\bauthor{\bsnm{Sorkine-Hornung}, \binits{O.}}:
\batitle{Autocuts: simultaneous distortion and cut optimization for uv mapping}.
\bjtitle{ACM Transactions on Graphics}
\bvolume{36}(\bissue{6}),
\bfpage{1}--\blpage{11}
(\byear{2017})
\end{barticle}
\endbibitem

\bibitem[\protect\citeauthoryear{Srinivasan et~al.}{2024}]{srinivasan2024nuvo}
\begin{bchapter}
\bauthor{\bsnm{Srinivasan}, \binits{P.P.}},
\bauthor{\bsnm{Garbin}, \binits{S.J.}},
\bauthor{\bsnm{Verbin}, \binits{D.}},
\bauthor{\bsnm{Barron}, \binits{J.T.}},
\bauthor{\bsnm{Mildenhall}, \binits{B.}}:
\bctitle{Nuvo: Neural uv mapping for unruly 3d representations}.
In: \bbtitle{Proceedings of the European Conference on Computer Vision},
pp. \bfpage{18}--\blpage{34}
(\byear{2024})
\end{bchapter}
\endbibitem

\bibitem[\protect\citeauthoryear{Majdisova and Skala}{2017}]{majdisova2017radial}
\begin{barticle}
\bauthor{\bsnm{Majdisova}, \binits{Z.}},
\bauthor{\bsnm{Skala}, \binits{V.}}:
\batitle{Radial basis function approximations: comparison and applications}.
\bjtitle{Applied Mathematical Modelling}
\bvolume{51},
\bfpage{728}--\blpage{743}
(\byear{2017})
\end{barticle}
\endbibitem

\bibitem[\protect\citeauthoryear{Naiman et~al.}{2017}]{naiman2017houdini}
\begin{barticle}
\bauthor{\bsnm{Naiman}, \binits{J.}},
\bauthor{\bsnm{Borkiewicz}, \binits{K.}},
\bauthor{\bsnm{Christensen}, \binits{A.}}:
\batitle{Houdini for astrophysical visualization}.
\bjtitle{Publications of the Astronomical Society of the Pacific}
\bvolume{129}(\bissue{975}),
\bfpage{058008}
(\byear{2017})
\end{barticle}
\endbibitem

\bibitem[\protect\citeauthoryear{Borkiewicz et~al.}{2019}]{borkiewicz2019cinematic}
\begin{barticle}
\bauthor{\bsnm{Borkiewicz}, \binits{K.}},
\bauthor{\bsnm{Naiman}, \binits{J.P.}},
\bauthor{\bsnm{Lai}, \binits{H.}}:
\batitle{Cinematic visualization of multiresolution data: Ytini for adaptive mesh refinement in houdini}.
\bjtitle{The Astronomical Journal}
\bvolume{158}(\bissue{1}),
\bfpage{10}
(\byear{2019})
\end{barticle}
\endbibitem

\bibitem[\protect\citeauthoryear{{SideFX}}{2023}]{SideFX2023Houdini}
\begin{botherref}
\oauthor{\bsnm{{SideFX}}}:
Houdini FX,
Toronto, Canada.
Software
(2023).
\url{https://www.sidefx.com}
\end{botherref}
\endbibitem

\bibitem[\protect\citeauthoryear{Ke et~al.}{2022}]{Harmonizer}
\begin{bchapter}
\bauthor{\bsnm{Ke}, \binits{Z.}},
\bauthor{\bsnm{Sun}, \binits{C.}},
\bauthor{\bsnm{Zhu}, \binits{L.}},
\bauthor{\bsnm{Xu}, \binits{K.}},
\bauthor{\bsnm{Lau}, \binits{R.W.H.}}:
\bctitle{Harmonizer: Learning to perform white-box image and video harmonization}.
In: \bbtitle{Proceedings of the European Conference on Computer Vision}
(\byear{2022})
\end{bchapter}
\endbibitem

\bibitem[\protect\citeauthoryear{Esser et~al.}{2024}]{esser2024scaling}
\begin{bchapter}
\bauthor{\bsnm{Esser}, \binits{P.}},
\bauthor{\bsnm{Kulal}, \binits{S.}},
\bauthor{\bsnm{Blattmann}, \binits{A.}},
\bauthor{\bsnm{Entezari}, \binits{R.}},
\bauthor{\bsnm{M{\"u}ller}, \binits{J.}},
\bauthor{\bsnm{Saini}, \binits{H.}},
\bauthor{\bsnm{Levi}, \binits{Y.}},
\bauthor{\bsnm{Lorenz}, \binits{D.}},
\bauthor{\bsnm{Sauer}, \binits{A.}},
\bauthor{\bsnm{Boesel}, \binits{F.}}, \betal:
\bctitle{Scaling rectified flow transformers for high-resolution image synthesis}.
In: \bbtitle{Proceedings of the International Conference on Machine Learning},
pp. \bfpage{12606}--\blpage{12633}
(\byear{2024})
\end{bchapter}
\endbibitem

\bibitem[\protect\citeauthoryear{Zhang et~al.}{2023}]{zhang2023adding}
\begin{bchapter}
\bauthor{\bsnm{Zhang}, \binits{L.}},
\bauthor{\bsnm{Rao}, \binits{A.}},
\bauthor{\bsnm{Agrawala}, \binits{M.}}:
\bctitle{Adding conditional control to text-to-image diffusion models}.
In: \bbtitle{Proceedings of the IEEE/CVF International Conference on Computer Vision},
pp. \bfpage{3836}--\blpage{3847}
(\byear{2023})
\end{bchapter}
\endbibitem

\bibitem[\protect\citeauthoryear{Betker et~al.}{2023}]{betker2023improving}
\begin{barticle}
\bauthor{\bsnm{Betker}, \binits{J.}},
\bauthor{\bsnm{Goh}, \binits{G.}},
\bauthor{\bsnm{Jing}, \binits{L.}},
\bauthor{\bsnm{Brooks}, \binits{T.}},
\bauthor{\bsnm{Wang}, \binits{J.}},
\bauthor{\bsnm{Li}, \binits{L.}},
\bauthor{\bsnm{Ouyang}, \binits{L.}},
\bauthor{\bsnm{Zhuang}, \binits{J.}},
\bauthor{\bsnm{Lee}, \binits{J.}},
\bauthor{\bsnm{Guo}, \binits{Y.}}, \betal:
\batitle{Improving image generation with better captions}.
\bjtitle{Computer Science. https://cdn. openai. com/papers/dall-e-3. pdf}
\bvolume{2}(\bissue{3}),
\bfpage{8}
(\byear{2023})
\end{barticle}
\endbibitem

\bibitem[\protect\citeauthoryear{Comanici et~al.}{2025}]{comanici2025gemini}
\begin{botherref}
\oauthor{\bsnm{Comanici}, \binits{G.}},
\oauthor{\bsnm{Bieber}, \binits{E.}},
\oauthor{\bsnm{Schaekermann}, \binits{M.}},
\oauthor{\bsnm{Pasupat}, \binits{I.}},
\oauthor{\bsnm{Sachdeva}, \binits{N.}},
\oauthor{\bsnm{Dhillon}, \binits{I.}},
\oauthor{\bsnm{Blistein}, \binits{M.}},
\oauthor{\bsnm{Ram}, \binits{O.}},
\oauthor{\bsnm{Zhang}, \binits{D.}},
\oauthor{\bsnm{Rosen}, \binits{E.}}, et al.:
Gemini 2.5: Pushing the frontier with advanced reasoning, multimodality, long context, and next generation agentic capabilities.
Preprint at https://doi.org/10.48550/arXiv.2507.06261
(2025)
\end{botherref}
\endbibitem

\bibitem[\protect\citeauthoryear{Harihar et~al.}{2019}]{harihar2019tigerrajaji}
\begin{botherref}
\oauthor{\bsnm{Harihar}, \binits{A.}},
\oauthor{\bsnm{Pandav}, \binits{B.}},
\oauthor{\bsnm{Hussein}, \binits{I.}}:
Camera trap database of Tiger from Rajaji National Park, Uttarakhand.
Wildlife Institute of India.
Accessed via GBIF.org on 2025-08-04
(2019)
\end{botherref}
\endbibitem

\bibitem[\protect\citeauthoryear{Han et~al.}{2023}]{han2023synergistic}
\begin{botherref}
\oauthor{\bsnm{Han}, \binits{B.A.}},
\oauthor{\bsnm{Varshney}, \binits{K.R.}},
\oauthor{\bsnm{LaDeau}, \binits{S.}},
\oauthor{\bsnm{Subramaniam}, \binits{A.}},
\oauthor{\bsnm{Weathers}, \binits{K.C.}},
\oauthor{\bsnm{Zwart}, \binits{J.}}:
A synergistic future for ai and ecology.
Proceedings of the National Academy of Sciences
\textbf{120}(38)
(2023)
\end{botherref}
\endbibitem

\bibitem[\protect\citeauthoryear{Kodi et~al.}{2024}]{kodi2024ghostbusting}
\begin{barticle}
\bauthor{\bsnm{Kodi}, \binits{A.R.}},
\bauthor{\bsnm{Howard}, \binits{J.}},
\bauthor{\bsnm{Borchers}, \binits{D.L.}},
\bauthor{\bsnm{Worthington}, \binits{H.}},
\bauthor{\bsnm{Alexander}, \binits{J.S.}},
\bauthor{\bsnm{Lkhagvajav}, \binits{P.}},
\bauthor{\bsnm{Bayandonoi}, \binits{G.}},
\bauthor{\bsnm{Ochirjav}, \binits{M.}},
\bauthor{\bsnm{Erdenebaatar}, \binits{S.}},
\bauthor{\bsnm{Byambasuren}, \binits{C.}}, \betal:
\batitle{Ghostbusting—reducing bias due to identification errors in spatial capture-recapture histories}.
\bjtitle{Methods in Ecology and Evolution}
\bvolume{15}(\bissue{6}),
\bfpage{1060}--\blpage{1070}
(\byear{2024})
\end{barticle}
\endbibitem

\bibitem[\protect\citeauthoryear{Cong et~al.}{2020}]{cong2020dovenet}
\begin{bchapter}
\bauthor{\bsnm{Cong}, \binits{W.}},
\bauthor{\bsnm{Zhang}, \binits{J.}},
\bauthor{\bsnm{Niu}, \binits{L.}},
\bauthor{\bsnm{Liu}, \binits{L.}},
\bauthor{\bsnm{Ling}, \binits{Z.}},
\bauthor{\bsnm{Li}, \binits{W.}},
\bauthor{\bsnm{Zhang}, \binits{L.}}:
\bctitle{Dovenet: Deep image harmonization via domain verification}.
In: \bbtitle{Proceedings of the IEEE/CVF Conference on Computer Vision and Pattern Recognition},
pp. \bfpage{8394}--\blpage{8403}
(\byear{2020})
\end{bchapter}
\endbibitem

\bibitem[\protect\citeauthoryear{Wang}{2022}]{wang2022interpolation}
\begin{bchapter}
\bauthor{\bsnm{Wang}, \binits{X.}}:
\bctitle{Interpolation and sharpening for image upsampling}.
In: \bbtitle{Proceedings of the International Conference on Computer Graphics, Image and Virtualization},
pp. \bfpage{73}--\blpage{77}
(\byear{2022})
\end{bchapter}
\endbibitem

\bibitem[\protect\citeauthoryear{Radford et~al.}{2021}]{radford2021learning}
\begin{bchapter}
\bauthor{\bsnm{Radford}, \binits{A.}},
\bauthor{\bsnm{Kim}, \binits{J.W.}},
\bauthor{\bsnm{Hallacy}, \binits{C.}},
\bauthor{\bsnm{Ramesh}, \binits{A.}},
\bauthor{\bsnm{Goh}, \binits{G.}},
\bauthor{\bsnm{Agarwal}, \binits{S.}},
\bauthor{\bsnm{Sastry}, \binits{G.}},
\bauthor{\bsnm{Askell}, \binits{A.}},
\bauthor{\bsnm{Mishkin}, \binits{P.}},
\bauthor{\bsnm{Clark}, \binits{J.}}, \betal:
\bctitle{Learning transferable visual models from natural language supervision}.
In: \bbtitle{Proceedings of the International Conference on Machine Learning},
pp. \bfpage{8748}--\blpage{8763}
(\byear{2021})
\end{bchapter}
\endbibitem

\bibitem[\protect\citeauthoryear{Jiang and Ye}{2023}]{cvpr23crossmodal}
\begin{bchapter}
\bauthor{\bsnm{Jiang}, \binits{D.}},
\bauthor{\bsnm{Ye}, \binits{M.}}:
\bctitle{Cross-modal implicit relation reasoning and aligning for text-to-image person retrieval}.
In: \bbtitle{Proceedings of the IEEE/CVF Conference on Computer Vision and Pattern Recognition}
(\byear{2023})
\end{bchapter}
\endbibitem

\bibitem[\protect\citeauthoryear{Dosovitskiy et~al.}{2021}]{dosovitskiy2021an}
\begin{bchapter}
\bauthor{\bsnm{Dosovitskiy}, \binits{A.}},
\bauthor{\bsnm{Beyer}, \binits{L.}},
\bauthor{\bsnm{Kolesnikov}, \binits{A.}},
\bauthor{\bsnm{Weissenborn}, \binits{D.}},
\bauthor{\bsnm{Zhai}, \binits{X.}},
\bauthor{\bsnm{Unterthiner}, \binits{T.}},
\bauthor{\bsnm{Dehghani}, \binits{M.}},
\bauthor{\bsnm{Minderer}, \binits{M.}},
\bauthor{\bsnm{Heigold}, \binits{G.}},
\bauthor{\bsnm{Gelly}, \binits{S.}},
\bauthor{\bsnm{Uszkoreit}, \binits{J.}},
\bauthor{\bsnm{Houlsby}, \binits{N.}}:
\bctitle{An image is worth 16x16 words: Transformers for image recognition at scale}.
In: \bbtitle{Preceedings of the International Conference on Learning Representations}
(\byear{2021})
\end{bchapter}
\endbibitem

\end{thebibliography}
\clearpage
\section{Methods}\label{sec4}

\begin{figure}[t]
\centering
\includegraphics[width=\textwidth]{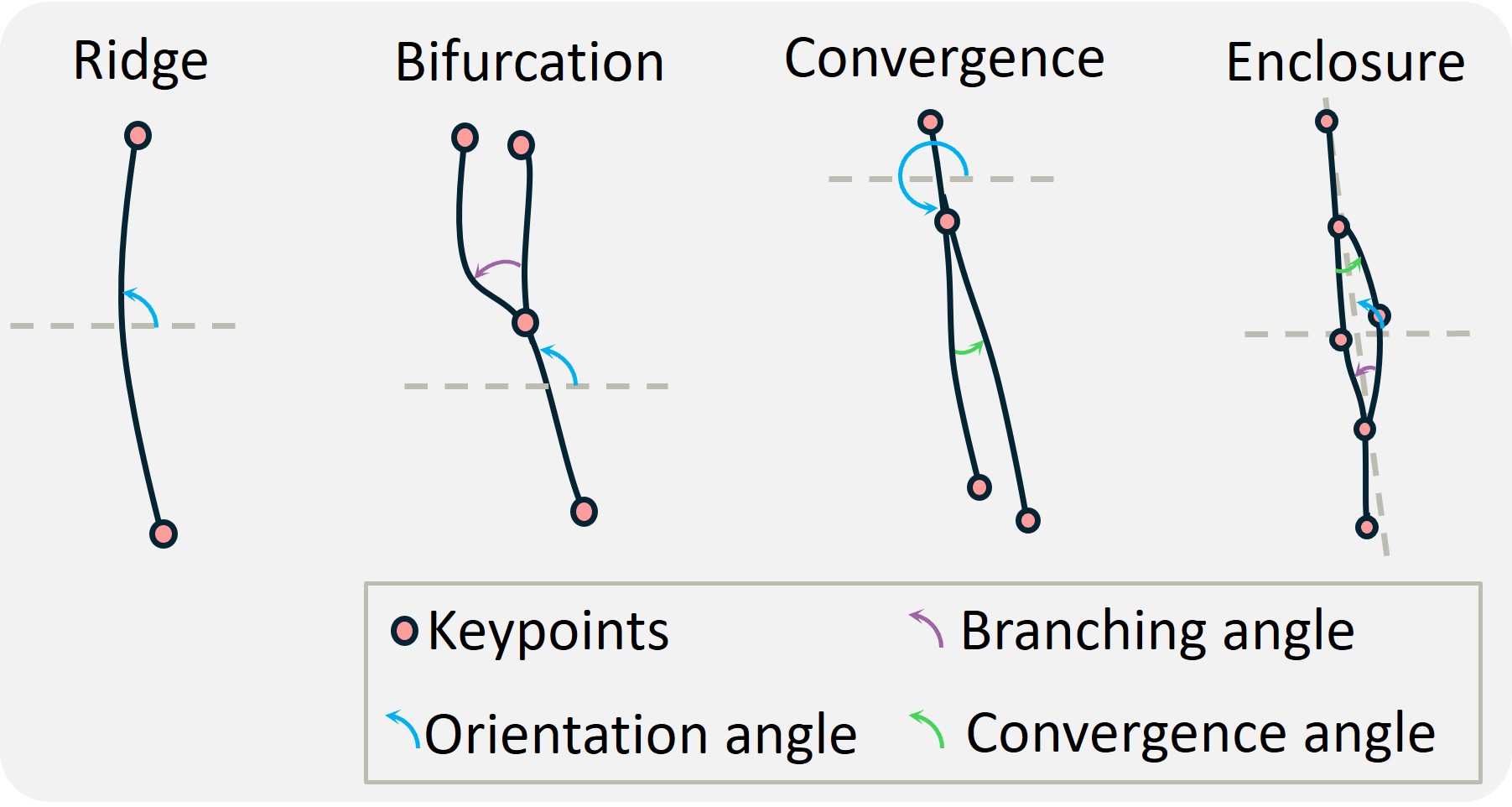}
\caption{{\textbf{Topological definition and parametrisation of four common dermatoglyphic structures.} Adapted from forensic taxonomies, schematics illustrate the spatial arrangement of keypoints (pink dots) and angular vectors (arrows) for four common structural dermatoglyphic types. These geometric definitions underpin the construction of an augmented minutiae library, facilitating the synthesis of individual-specific patterns.}
}\label{minutiae_detail}
\end{figure}

\subsection{{Minutiae Definitions and Stripe Library Construction}}\label{method_minutiae_augmentation}

{\textbf{Structural Minutiae Definitions.} Our system of ID description is anchored in four topological minutiae types found in tiger patterns; these are akin to oriented fingerprint details in forensic science. Each minutiae is defined based on fixed, directed arrangements of keypoints and their connections which, by adding angles and distance information, can be rendered as particular geometric stripe arrangements~(see Fig.~\ref{minutiae_detail}).
As in forensics, a \textit{ridge} is defined as a continuous segment~(i.e. a stripe) with a single orientation angle connecting two endpoints. {A \textit{bifurcation} and a  \textit{convergence}} represent branchings of a primary stripe into two diverging paths. {Each is parametrised by four keypoints overall, a central junction and three branch endpoints,} and two angles, the original ridge orientation angle and the branching angle. A \textit{convergence} can, of course, be understood as a vertically flipped bifurcation whose orientation angle lies within the range of $180^\circ$ to $360^\circ$ (see Fig \ref{minutiae_detail} blue arrows). Together, these three directed minutiae types allow for the synthesis of arbitrary, oriented stripe topologies including those on tigers. Following forensic science to improve descriptiveness, we add an \textit{enclosure} minutiae type which is a compound structure formed by a bifurcation followed by a close-by convergence in a closed loop defined by six keypoints: two terminals, one bifurcation point, one convergence point, and two arc apexes,  and three angles: branching angle, convergence angle, and global orientation angle. Fig.~\ref{minutiae_detail} depicts full details.}

{\textbf{Stripe Library with Typed Minutiae Annotations.} We utilise minutiae definitions for constructing our \textit{stripe} library (database) that forms the foundation of visual-textual co-synthesis~(as depicted before in Fig.~\ref{statistical pattern synthesis}). To do this as close as possible to natural stripe geometries, we sample large sets of stripe arrangements from real tiger images and classify them based on minutiae contained. We then apply realistic geometry transforms to augment the data, including random translation, affine transformation, local non-linear distortion, rotation, and scale adjustment. During the affine transformation stage, random perturbations are applied to each keypoint $(x_i, y_i)$ to obtain new coordinates $(x'_i, y'_i)$. A homography matrix $H$ was computed such that}
\begin{equation}
\begin{bmatrix} x'_i \\ y'_i \\ 1 \end{bmatrix} \sim H \begin{bmatrix} x_i \\ y_i \\ 1 \end{bmatrix} .
\end{equation}
{The stripe images themselves were transformed using inverse mapping, that is for any target pixel $(x', y')$, its corresponding position in the augmented source image was computed as}
\begin{equation}
I'(x', y') = I\left(H^{-1} \begin{bmatrix} x' \\ y' \\ 1 \end{bmatrix} \right),
\end{equation}
{where $I$ and $I'$ denote the images before and after transformation, respectively. For local non-linear distortion augmentation, we selected a starting point $p_i = (x_i, y_i)$ and a target point $p_j = (x_j, y_j)$, and defined a distortion strength coefficient $k_0$ and influence radius $r$ to enable local deformation. Any pixel $p = (x, y)$ was transformed to}
\begin{equation}
\begin{aligned}
p' &= p - \rho (p_j - p_i)\left(1 - \frac{k_1}{r}\right), \\[6pt]
\text{where}\quad 
k_1 &= \sqrt{(x - x_i)^2 + (y - y_i)^2}, \\[6pt]
\text{and}\quad 
\rho &= 
\left(
\frac{r^2 - \|p - p_i\|^2}
     {r^2 - \|p - p_i\|^2 + k_0 \cdot \|p_j - p_i\|^2}
\right)^2 .
\end{aligned}
\end{equation}
{The random translation magnitude was uniformly sampled between 0 and 15 pixels, the local distortion radius was set to 50 pixels, the strength parameter $k_0$ was set to 200, the rotation angle was sampled from $[-15^\circ, +15^\circ]$, and the scaling factor ranged from 0.8 to 1.2.}



{\subsection{Implementation Details of Virtual Animal Modelling and Rendering}}\label{method_RBF_and_rendering_detail}
{\textbf{RBF-based Distortion-compensated Projection.} During the projection from UV coat pattern space to the 3D animal mesh surface~(shown earlier in Fig.~\ref{rbf and skeleton} (a)), we use radial basis function to compensate for distortions caused by deformation. Specifically, we select a set of points in UV space $\{\mathbf{p}_i = (x_i, y_i)\}_{i=1}^N$ and their corresponding points in 3D space. The horizontal and vertical deformations are respectively fitted by $f_x(x, y)$ and $f_y(x, y)$ as}
\begin{equation}
\begin{aligned}
f_x(x, y) &= \sum_{i=1}^{N} w^{(x)}_i \,\phi\!\left( \left\lVert (x, y) - (x_i, y_i) \right\rVert \right), \\[6pt]
\text{and}\quad
f_y(x, y) &= \sum_{i=1}^{N} w^{(y)}_i \,\phi\!\left( \left\lVert (x, y) - (x_i, y_i) \right\rVert \right).
\end{aligned}
\end{equation}
{where $\phi(r) = \sqrt{r^2 + \varepsilon^2}$ denotes the multiquadric kernel function, $\lVert \cdot \rVert$ is the Euclidean distance, and the weight parameters $w_i^{(x)}$ and $w_i^{(y)}$ are obtained by solving the associated linear system.}

{\textbf{Parameterisation of Hair Synthesis Process.} Pelage modelling, as visualised earlier in Fig.~\ref{hair}, was completed in Houdini 20.0~\cite{SideFX2023Houdini} with settings for hair guide curves as: density of 20,000, relax iterations set to 3, influence radius of 0.0322, influence decay of 2, and max guide angle of $90^\circ$.} 
{The random variation in hair length was configured differently depending on the hair type. For short and medium hair, the \texttt{Set} mode was used, with random length ranges of 0.01–0.015 and 0.015–0.02, respectively. For long hair, the \texttt{Multiply} mode was used, with a random scaling factor ranging from 1.0 to 1.5.}
{For the \texttt{Frizz} effect, the frequency was uniformly set to 40, and the amplitudes were 0.0092 for short hair, 0.0011 for medium hair, and 0.0185 for long hair. The \texttt{Clump} effect was applied only to medium and long hair: for medium hair, the blend factor was 1.0 and the clump size was 0.141; for long hair, the blend factor was 0.7 and the clump size was 0.018.}

{\textbf{Virtual Scene Lighting.} To simulate natural lighting conditions, Houdini is set to incorporate three types of light sources: environment light, sunlight, and skylight. The environment light source uses an 8K HDRI (high dynamic range imaging) captured from a real nature reserve, with light intensity set to 5 and exposure set to 0 (see Fig \ref{render}). The sunlight parameters are: white light, intensity = 5, exposure = 0, cone angle = 45, cone delta = 10, and cone rolloff = 1. The skylight is configured with intensity = 0.6 and exposure = 0.}

{\textbf{Image Rendering Environment.} Image rendering itself is performed using the Karma XPU renderer (with CPU and GPU hybrid acceleration). The hardware platform includes an Intel Core i9-13980HX CPU and an NVIDIA RTX 4080 GPU. The rendering resolution is set to $640 \times 360$. Path traced samples are set to 128, light sampling quality to 1, volume step rate to 0.25, diffuse limit to 1, and reflection limit to 4. In the output images, the animal foreground retains a transparent background.}

{\textbf{Background and Image Fusion Details.} Background images for virtual camera trap imagery are selected from animal-free regions of 79 real camera trap images, covering various seasons and lighting conditions. Foregrounds and backgrounds are fused using the image harmonisation model 
\cite{Harmonizer}, which is pre-trained on the iHarmony4 \cite{cong2020dovenet} dataset. To match the quality of real camera trap images from our datasets, the synthesised visuals undergo effective downsampling (with a scaling factor of 0.7 using Lanczos method\cite{wang2022interpolation}). In addition, 20\% of the images in the dataset are augmented with Gaussian noise (intensity = 0.07), and 10\% are augmented with motion blur (blur radius = 3) to further simulate effects in real camera trap data distributions. Gradient transparency~(at transparency strength = 1) was added to the animal's feet to achieve integration with the ground.}

\vspace{10pt}
\subsection{{Datasets}}\label{data}
{\textbf{Overview.} Tiger datasets used for this case study cover  a real-world camera trap dataset and a synthetic dataset. In both cases, according to standard dataset construction protocols, the two sides of the same animal are treated as separate individuals.}

\begin{figure}[th]
\centering
\includegraphics[width=\textwidth]{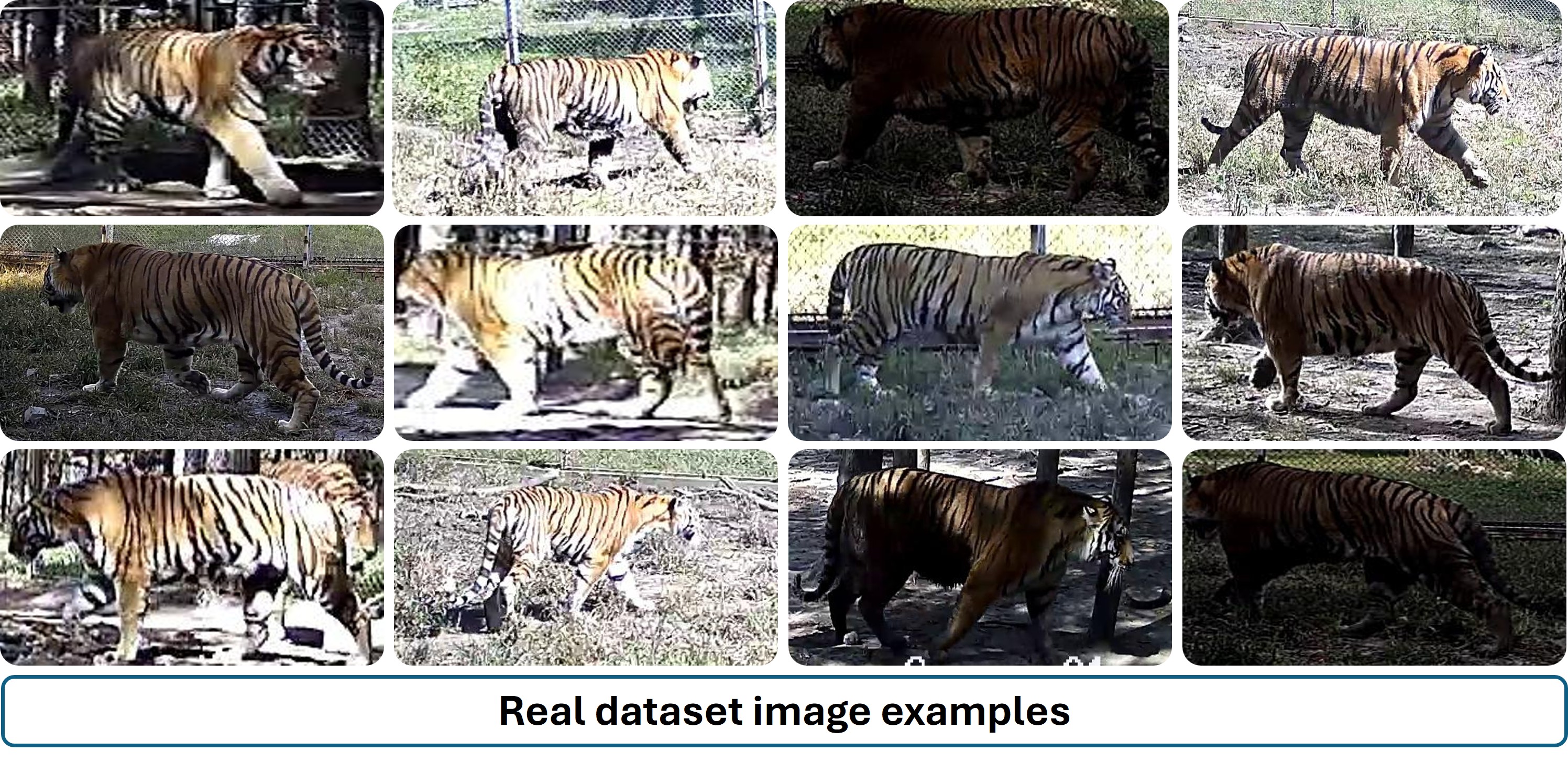}
\caption{\textbf{Examples of Real-world Data Samples.} 
The authentic images shown are sourced from camera and cropped using object detection bounding boxes. This real-world data serves as a benchmark for our synthesis and rendering pipeline, which highly simulates the authentic distribution of viewing angles, animal poses, illumination conditions, and image quality. To address the relatively monolithic backgrounds present in the real-world dataset—a result of limited animal individuals and camera sites—our synthesis pipeline introduces greater background and illumination diversity. This strategy mitigates model bias, preventing the learning of spurious background correlations and forcing the model to learn intrinsic animal features, thereby enhancing generalisation.}\label{real data}
\end{figure}

{\textbf{Real-world Dataset.} This contains 185 individuals and a total of 3{,}355 images~(see Fig.~\ref{real data}), each cropped and paired with manually annotated, ACE-based dermatoglyphic descriptions of the coat pattern features. The images are sourced from the ATRW dataset~\cite{li2020atrw} and the Camera Trap Database of Tigers~\cite{harihar2019tigerrajaji} from Rajaji National Park, Uttarakhand, India. The dataset is split into a training set~(165 individuals) and a test set~(20 individuals) for cross-model Re-ID, using a random assignment strategy.}

{\textbf{Synthetic Data.} This covers 2{,}000 synthesised individuals across 24{,}000 images~(exemplified in Fig.~\ref{render}), with 12 images per virtual tiger under varying viewpoints and pose. Each image is accompanied by a corresponding, generated description of its stripe pattern structure. In ablation experiments, the full synthetic dataset is divided into subsets of 200 individuals each. Dermatoglyphic coat descriptions in the synthetic data are randomly selected based on multiple anchor permutations.}

\subsection{{AI Model Architectures and Loss Functions}}\label{DL}
{\textbf{Visual Network Architecture.} In this study, we adopt the CLIP model~\cite{radford2021learning,cvpr23crossmodal} provided by OpenAI as the baseline, where the image encoder uses the ViT architecture~\cite{dosovitskiy2021an}. For pre-processing, input images are uniformly resized and divided into non-overlapping $16 \times 16$ patches. Each patch is then flattened and linearly projected, with a learnable [CLS] token prepended to the sequence, and corresponding positional encodings added. The sequence is then passed through 12 layers of Transformer encoders, each consisting of LayerNorm, multi-head self-attention, residual connections, feed-forward networks, and a second residual path~{shown previously in Fig.~\ref{network} (b)}. The final output corresponding to the [CLS] token is extracted, projected by a linear layer, and L2-normalised to serve as the image embedding vector aligned with text features.}

{\textbf{Language Network Architecture.} The text encoder also follows the Transformer-based CLIP structure. Input texts are first tokenised using Byte Pair Encoding~(BPE) into variable-length token sequences (up to a maximum of 77 tokens), and mapped through a learnable embedding layer. Two special [SOS] and [EOS] tokens are appended to the start and the end of each sequence to aggregate sentence-level semantics. Positional embeddings are added to the entire sequence, which is then input into a 12-layer Transformer encoder for feature extraction, structured identically to the image encoder. Finally, the output at the position of the [EOS] token is taken as the sentence-level semantic representation, projected through a linear layer, and L2-normalised to serve as the text embedding vector. Fig.~\ref{network} visualises these details.}

{\textbf{Re-identification Loss.} To effectively guide the model in learning discriminative embedding representations, we introduce a variant of the triplet loss based on hard example mining. Let the batch of global features be denoted as $\mathbf{X} = \{ \mathbf{x}_1, \mathbf{x}_2, \dots, \mathbf{x}_N \}$ with corresponding labels $\mathbf{y}$. First, we compute the pairwise Euclidean distance matrix $\mathbf{D} \in \mathbb{R}^{N \times N}$, where}
\begin{equation}
D_{ij} = \| \mathbf{x}_i - \mathbf{x}_j \|_2 = \sqrt{ \| \mathbf{x}_i \|^2 + \| \mathbf{x}_j \|^2 - 2 \mathbf{x}_i^\top \mathbf{x}_j }.
\end{equation}
{For numerical stability, all features can optionally be L2-normalised:}
\begin{equation}
\mathbf{x}_i \leftarrow \frac{\mathbf{x}_i}{\| \mathbf{x}_i \|_2 + \epsilon}.
\end{equation}
{When constructing triplets $(\mathbf{x}_a, \mathbf{x}_p, \mathbf{x}_n)$, for each anchor sample $\mathbf{x}_a$, the hardest positive sample $\mathbf{x}_p$ (with the same label) and the hardest negative sample $\mathbf{x}_n$ (with a different label) are selected as:}
\begin{equation}
\text{dist}_{ap} = \max_{j: y_j = y_a} D_{aj}, \quad \text{dist}_{an} = \min_{k: y_k \neq y_a} D_{ak}.
\end{equation}
{To dynamically adjust the training difficulty, we introduce a difficulty factor $\gamma \in [0, 1)$ to weight the positive and negative distances:}
\begin{equation}
\tilde{d}_{ap} = (1 + \gamma) \cdot \text{dist}_{ap}, \quad \tilde{d}_{an} = (1 - \gamma) \cdot \text{dist}_{an}.
\end{equation}
{The final loss function for the Re-ID experiments can then be defined in a margin-based form as}
\begin{equation}
\mathcal{L} = \max(0, \tilde{d}_{ap} - \tilde{d}_{an} + m).
\end{equation}

{\textbf{Loss for Textual-to-Visual Retrieval.} To effectively align image and text features in the retrieval task while enhancing the model's ability to distinguish individual instances, we jointly use the Image-Text Contrastive Loss~(ITC) and the Instance Discrimination Loss~(ID) for training. The combined total loss is the unweighted sum}
\begin{equation}
L_{\text{total}} = L_{\text{ITC}} + L_{\text{ID}}
\end{equation}
{The ITC loss is essentially an implementation of InfoNCE in a cross-modal setting aiming to maximise the similarity between matched image-text pairs while minimising the similarity between mismatched pairs. Let the image features be $\mathbf{I} = \{I_i\}_{i=1}^{N}$ and the text features be $\mathbf{T} = \{T_i\}_{i=1}^{N}$. After normalisation, the cosine similarity matrix is computed as}
\begin{equation}
\mathbf{S}_{I \rightarrow T}(i,j) = s \cdot \frac{I_i^\top T_j}{\|I_i\|\|T_j\|}
\end{equation}
{where $s$ is a learnable temperature scaling factor~(using logit scale of 50). The loss is computed symmetrically in both image-to-text and text-to-image directions. The expanded form thus equates to}
\begin{equation}
L_{\text{ITC}} = -\frac{1}{2N}\left[\sum_{i=1}^{N}\log\frac{\exp(\mathbf{S}_{I \rightarrow T}(i,i))}{\sum_{j=1}^{N}\exp(\mathbf{S}_{I \rightarrow T}(i,j))} + \sum_{i=1}^{N}\log\frac{\exp(\mathbf{S}_{T \rightarrow I}(i,i))}{\sum_{j=1}^{N}\exp(\mathbf{S}_{T \rightarrow I}(i,j))}\right]
\end{equation}
{To further enhance the model's ability to separate individuals, we introduce an ID loss, which optimises the logits output from both the image and text branches in a supervised classification manner. Let the image prediction logits be $\mathbf{z}_I \in \mathbb{R}^{N \times C}$ and the text prediction logits be $\mathbf{z}_T \in \mathbb{R}^{N \times C}$, where~$C$ is the total number of classes and $y_i$ is the ground-truth label for the $i$-th sample. The ID loss can then be defined as}
\begin{equation}
L_{\text{ID}} = -\frac{1}{2N}\left[\sum_{i=1}^{N}\log\frac{\exp(\mathbf{z}_I(i,y_i))}{\sum_{c=1}^{C}\exp(\mathbf{z}_I(i,c))} + \sum_{i=1}^{N}\log\frac{\exp(\mathbf{z}_T(i,y_i))}{\sum_{c=1}^{C}\exp(\mathbf{z}_T(i,c))}\right]
\end{equation}

\subsection{{Experimental Environment}}\label{ES}

{All experiments were conducted on an HPC cluster based on the Slurm system. Training utilised NVIDIA GeForce RTX 2080 Ti, Tesla V100 (16GB and 32GB), NVIDIA RTX 3090, and NVIDIA A100 (80GB SXM4) GPUs.}

{\textbf{Re-identification Training Configuration.} The optimiser used is Adam, with an initial learning rate of 0.0003 and a bias learning rate multiplier of 2. The first 5 epochs adopt a linear learning rate warm-up strategy with a warm-up factor of 0.01. The regularisation parameters are set as: weight decay of 0.0005 and the same weight decay applied to bias terms.}

{\textbf{Retrieval Training Configuration.} Adam is also used as the optimiser, with an initial learning rate of $1 \times 10^{-5}$ and a bias learning rate multiplier of 2.0. The momentum is set to 0.9. The weight decay is $4 \times 10^{-5}$, and no weight decay is applied to bias terms. The warm-up phase lasts for 5 epochs with linear scheduling, and the warm-up factor is set to 0.1.}

\end{document}